\pdfoutput=1
\documentclass[letterpaper]{article} % DO NOT CHANGE THIS
\usepackage{aaai24}  % DO NOT CHANGE THIS
\usepackage{times}  % DO NOT CHANGE THIS
\usepackage{helvet}  % DO NOT CHANGE THIS
\usepackage{courier}  % DO NOT CHANGE THIS
\usepackage[hyphens]{url}  % DO NOT CHANGE THIS
\usepackage{graphicx} % DO NOT CHANGE THIS
\usepackage{natbib}  % DO NOT CHANGE THIS AND DO NOT ADD ANY OPTIONS TO IT
\usepackage{caption} % DO NOT CHANGE THIS AND DO NOT ADD ANY OPTIONS TO IT
\usepackage{epsfig}
\usepackage{amsmath}
\usepackage{amssymb}
\usepackage{multirow}
\usepackage{comment}
\usepackage{algorithm}
\usepackage{algpseudocode}
\usepackage{nicefrac}
\usepackage[table,xcdraw]{xcolor}
\usepackage{booktabs}
\usepackage{subcaption}
\usepackage{pifont}
\usepackage[T1]{fontenc}% http://ctan.org/pkg/fontenc
\usepackage[outline]{contour}% http://ctan.org/pkg/contour
\contourlength{0.1pt}
\contournumber{10}
\DeclareMathOperator*{\argmax}{arg\,max}
\DeclareMathOperator*{\argmin}{arg\,min}

\usepackage{pifont}
\usepackage[T1]{fontenc}
\usepackage{newfloat}
\usepackage{listings}
\DeclareCaptionStyle{ruled}{labelfont=normalfont,labelsep=colon,strut=off} % DO NOT CHANGE THIS
\floatstyle{ruled}
\newfloat{listing}{tb}{lst}{}
\floatname{listing}{Listing}
\title{Is it an i or an l: Test-time Adaptation of Text Line Recognition Models}
\author{Debapriya Tula, Sujoy Paul, Gagan Madan, Peter Garst, Reeve Ingle, Gaurav Aggarwal}
\affiliations{Google Research}

\usepackage{bibentry}
\begin{document}

\maketitle

\begin{abstract}
Recognizing text lines from images is a challenging problem, especially for handwritten documents due to large variations in writing styles. While text line recognition models are generally trained on large corpora of real and synthetic data, such models can still make frequent mistakes if the handwriting is inscrutable or the image acquisition process adds corruptions, such as noise, blur, compression, etc. Writing style is generally quite consistent for an individual, which can be leveraged to correct mistakes made by such models. Motivated by this, we introduce the problem of adapting text line recognition models during test time. We focus on a challenging and realistic setting where, given only a single test image consisting of multiple text lines, the task is to adapt the model such that it performs better on the image, without any labels. We propose an iterative self-training approach that uses feedback from the language model to update the optical model, with confident self-labels in each iteration. The confidence measure is based on an augmentation mechanism that evaluates the divergence of the model’s prediction in a local region. We perform rigorous evaluation of our method on several benchmark datasets as well as their corrupted versions. Experimental results on multiple datasets spanning multiple scripts show that the proposed adaptation method offers an absolute improvement of up to 8\% in character error rate with just a few iterations of self-training at test time.
\end{abstract}

\section{Introduction}

% Pointers:
% 1) TTA problem statement - lit survey, not specific to OCR, single image to be highlighted
% 2) How TTA can be used in HTR 
% 3) similar things in lit (Meta HTR, other classification based ones, BN SITA TENT TTT MEMO)
% 4) How our method is different from existing methods

% Decoder detached from main model, possibility to use many decoders

Text line recognition \cite{Diaz2021RethinkingTL,li2021trocr} has been a challenging problem in the field of computer vision and machine learning for several decades. The task of recognizing handwritten text involves understanding and interpreting human handwriting - having a free flowing nature \cite{8954058} - in various languages and styles, making it a complex and multifaceted problem. Over the years, various sophisticated models \cite{Diaz2021RethinkingTL,ingle2019scalable,li2021trocr,breuel2013high,long2021scene}
%8954058, decouple, Fogel2020ScrabbleGANSV, seq2seqdomain, fillup, 1211511, 23114}
have been developed, which are trained on large corpora of labeled and synthetic data. However, recognizing handwriting data still remains a challenge due to large variations in styles across individuals. Additionally, the image acquisition process often adds corruptions such as blur, noise, compression, making recognition even more challenging. Deep learning models are often known to have generalization issues leading to significant drop in performance for distribution shifts \cite{hoffman2018cycada,tsai2018learning}, corruptions \cite{hendrycks2019benchmarking}, etc. This is no different for text line recognition. To solve this problem, we develop an algorithm to adapt an existing text line recognition model to specific writing styles during test time, given only a single test image. 
% To the best of our knowledge, this is the first of a kind effort to adapt text line recognition models based on test input for improved performance, without specifically training to identify writers.

\begin{figure}
    \centering
    \includegraphics[scale=0.4]{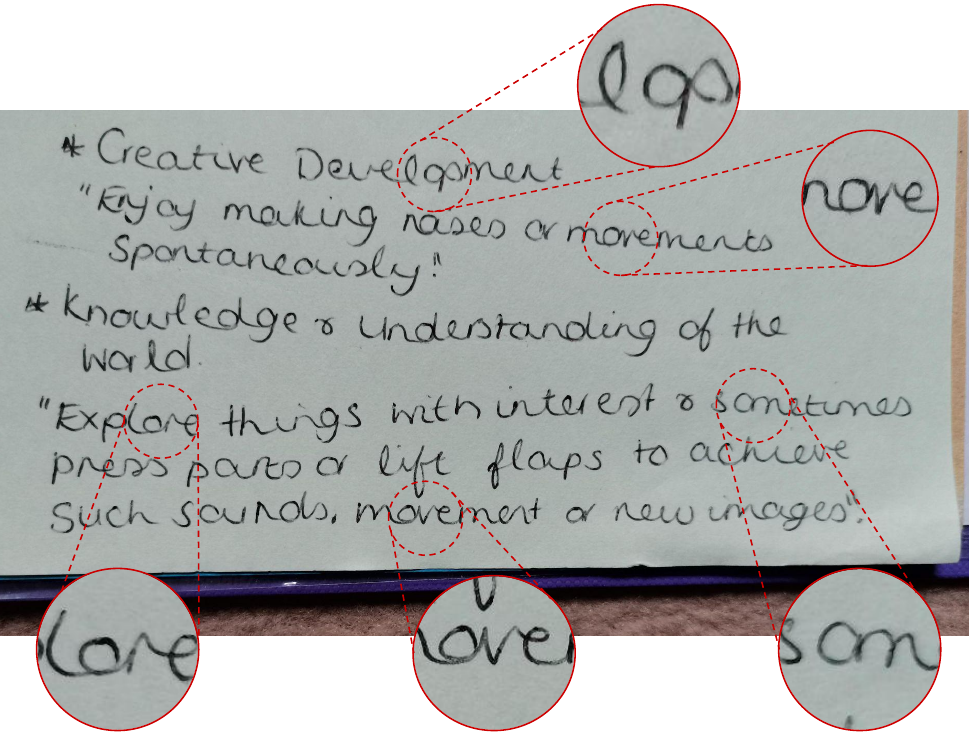}
    \caption{\small Sample image from the GNHK dataset \cite{goodnotes} showing how the letter "o" might look like "q" or "a" in handwritten text, which is the writer's style. We want our algorithm to adapt to such styles on the fly during test time.}
    \label{fig:intro}
    \vspace{-0.5 em}
\end{figure}

Unsupervised domain adaptation \cite{hoffman2018cycada,tsai2018learning} is a well-studied problem, where the task is to adapt a model trained on a labeled source dataset to a new target dataset using only unlabeled samples from the target. Such a paradigm may not align with real world settings where we are given only a single test instance for adaptation, without having access to any training data. Recently, there has been a pragmatic direction of work namely Test Time Adaptation (TTA) \cite{sun2020test,wang2021tent,sita}, where a model needs to be adapted on the fly, using only a few test samples, with no access to the training data. While these models need multiple test instances to show performance gains, we may not have multiple pages of handwriting for every test writer or style. Hence, we specifically look into the problem of single image test time adaptation for text line recognition models. 

\begin{table}[t]
			\caption{\small Comparison of different methods of adaptation. Our TTA setting considers a very challenging realistic scenario where the model has to be adapted to a single handwriting image, without access to any source data, target data or target labels.
			}
			\label{table:comp_rel}
			\centering
			\renewcommand{\arraystretch}{1.2}
	        \setlength{\tabcolsep}{4pt}
	        \resizebox{0.45\textwidth}{!}{
			\begin{tabular}{l|cc|c|ccc|c}
		    \toprule
		    \multirow{2}{*}{Method} & \multicolumn{2}{c}{Source data} & \multirow{2}{*}{Target Label} & \multicolumn{3}{c}{Target data} & Writer-specific\\
		    & Few & Many & & One & Few & Many & Information\\ 
		    \midrule
		    \cite{seq2seqdomain} & & \multirow{3}{*}{\checkmark} & \multirow{3}{*}{}  & & & \multirow{3}{*}{\checkmark} &\\
		     \cite{tang2022domain} &&&&&&& \\
		     \cite{syn2real} &&&&&&& \\
		    \midrule
		    \cite{metahtr} & \checkmark & & \checkmark & & \checkmark & & \checkmark \\
		    \midrule
		    \cite{kohut2023towards} & \checkmark & &  & & \checkmark & & \checkmark \\
		    \midrule
		    \cite{wang2022fast} & & \checkmark &  &  & &  & \checkmark  \\
		    \midrule
		    Ours & & &  & \checkmark & & \\
		   \bottomrule
	    \end{tabular}
	    }
\end{table}

Writing style generally varies significantly across individuals, and is often considered as the signature of an individual. For example, some writer's "i" may look like an "l", "n" may look like an "r", and so on (Figure~\ref{fig:intro}). However, instead of manually listing all such idiosyncrasies in handwriting across individuals, we would want to develop an algorithm that can automatically learn them on the fly. Typically, one individual's idiosyncrasies in writing style do not carry over to other individuals. Thus, we do not look into the online paradigm of updating models, but rather reset the model to the source model after every adaptation. This not only saves the model from diverging away, but also avoids any potential privacy concerns \cite{wang2022reconstructing}. 

TTA for image classification and semantic segmentation has been of recent interest in the literature. A few TTA methods \cite{sun19ttt, bartler2022mt3} require access to training data to learn from additional self-supervised losses which can then be optimized for the test instances. 
%A lot of methods have been explored to make models robust to data distribution shifts in test data. Early approaches to TTA like \cite{sun19ttt} and \cite{hendrycks2020augmix} require access to training data and augment it to withstand unforeseen corruptions during test time, or use specialised losses over the training data \cite{liang2020we}. 
%This shift in data distribution can also be induced by natural variations occurring in real world data, and hence may not be exactly replicated using standard augmentation techniques. 
Other methods like \cite{wang2021tent, memo} do not assume access to source data, but adapt to test data by minimising entropy over a batch where each batch has several data samples. On the contrary, we look into the problem of adapting to one writer style at a time, using only a single test instance. We also do not modify the training strategy, which would otherwise need access to the training data. Moreover, some of the metrics such as entropy used in these works, are non-trivial to compute for text line recognition models specifically for CTC decoder based models, as there can be many mappings to the same output string~\cite{graves2006connectionist}. 

Existing works on adaptation of text line recognition models require access to lots of unlabeled target data along with labeled source data \cite{seq2seqdomain,tang2022domain,Fogel2020ScrabbleGANSV,syn2real}. 
%to source data: labeled \cite{origaminet, supervised2}, partially labeled \cite{Fogel2020ScrabbleGANSV, seq2seqdomain, syn2real} or unlabeled \cite{seqclr}.
But this may not always be possible due to privacy/storage concerns entailing the source data. Other methods like \cite{metahtr,kohut2023towards} need a few labeled samples of the new writer to adapt the source model. Our method does not need any access to the source data, can be applied on any off-the-shelf text line recognizer, and does not need any labeled data during test-time for adaptation. To the best of our knowledge, TTA from a single handwritten image of a writer, without access to any source data, and without using any writer identification information during source model training, is a novel and more realistic setting which has not been explored yet. Table \ref{table:comp_rel} shows the comparison of different related works in the literature.

\begin{figure*}
    \centering
    \includegraphics[scale=0.35]{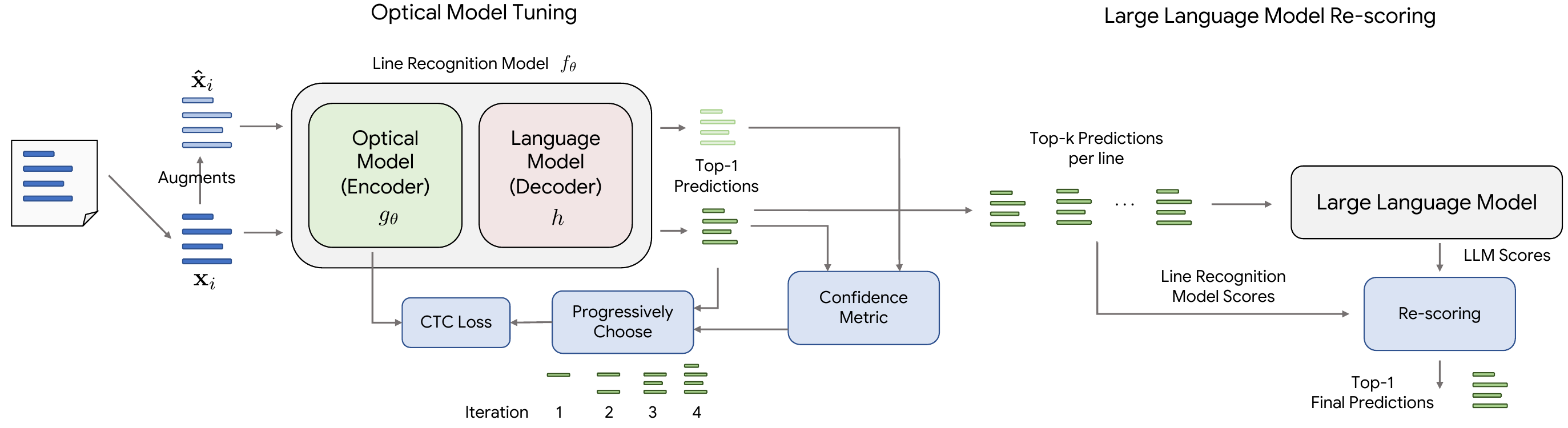}
    \caption{\small \textbf{Framework Overview.} Given an input image, we extract the lines from it. Then pass the original and the augmented version of the lines through the line recognition model which comprises of an optical and a language model. We then compare the outputs for the original and the augmented version of lines to get a confidence measure for each line. Based on this measure, we progressively choose lines and use their self-labels to compute the CTC loss and update the optical model. After updating the optical model in this way, we then extract top-k predictions for every line along with their scores. We then pass these top-k lines through a large language model to obtain the likelihood scores and combine it with the line recognition scores to re-score the top-k predictions and finally predict only the top-1.}
    \label{fig:framework}
\end{figure*}

Our TTA setting takes as input a single handwritten image of a writer containing a few lines of handwritten text in it. We look into a de-coupled text-line recognition model consisting of an encoder (optical part) and decoder (language model). Such a model allows us to plug-and-play the language model (LM) based on the domain at hand. Our algorithm exploits both the encoder and the decoder for adaptation. As shown in Figure \ref{fig:framework}, we first progressively update the optical model using the output from the LM decoder via a weighted CTC loss. The weights are obtained by judging the model's confidence in a local region around the input image. The iterative nature helps the algorithm to self-improve beyond the original model. We use a computationally efficient character n-gram language model within the loop to get superior quality pseudo-labels than just the optical model's prediction, which acts as an additional supervisory source. After updating the optical model, which more often looks at local context, we exploit longer context information via a Large Language Model (LLM). Specifically, we get the top-k predictions from the updated model using the beam search algorithm, and then re-rank them based on log-likelihood score of a large language model, which looks at the entire line. We show that both these steps are complementary to each other and offer a significant improvement in performance.

%The OCR model is then adapted iteratively over these lines. Our method involves an augmentation based strategy for selecting a fraction of confident lines among all lines of the image. The predictions over these confident lines then serve as pseudo labels over which the OCR model is to be trained on through backpropagation over iterations. The trained model is then used to choose a larger fraction of confident lines, on which the model is further trained on. This process is continued until all lines in the image are considered in the fraction. We try a few techniques to choose the fraction of confident lines, as well as to prevent our model from catastrophic forgetting during the adaptation phase. This iteratively trained model is adapted for one single test image at a time, and is reset to its initial state for each new test instance.

We perform experiments on five benchmark datasets: ICDAR2015-HTR \cite{icdar2015}, GNHK \cite{goodnotes}, IAM \cite{iam}, CVL \cite{kleber2013cvl} and KOHTD \cite{kohtd}. We also create corrupted versions of these datasets similar to \cite{hendrycks2019benchmarking} for image classification, to simulate corruptions which may occur during image acquisition. We perform rigorous qualitative and quantitative experiments, and ablation studies to show the efficacy of the proposed method. The key contributions of this paper are:

\begin{enumerate}
\itemsep0em
\item We introduce the problem of single image test time adaptation for the novel objective of adapting text line recognition to individual styles.
\item We develop an algorithm that uses LM in the loop to update the optical model, followed by refining the predictions by looking at longer context using an LLM.
\item Our algorithm shows significant performance improvements over baselines \cite{Diaz2021RethinkingTL} on multiple datasets.
\end{enumerate}

%-------------------------------------------------------------------------

\section{Related Work}

{\flushleft \textbf{Test-Time Adaptation.}} 
Test-Time Adaptation (TTA) has been recently studied in the literature for classification \cite{ wang2021tent,bartler2022mt3,sita} and segmentation tasks \cite{shin2022mm,sita}. These methods can be grouped into two broader categories. First, methods in which the training algorithm includes additional heads for self-supervised tasks \cite{sun2020test,prabhudesaitest,bartler2022mt3}. These additional losses are also optimized on the test images during test time. Self-supervised losses include rotation prediction \cite{sun2020test}, self-reconstruction \cite{prabhudesaitest}, student-teacher feature prediction \cite{bartler2022mt3}, etc. The hypothesis is that such losses not only regularize the model during training, but also improve the performance when optimized for test samples, given the gradients for such losses align with the gradients computed w.r.t. the actual labels of the test samples. The second group of methods, on the other hand, do not modify the training strategy and only update the source model using some pseudo-losses such as entropy \cite{memo, wang2021tent}, self-label cross-entropy \cite{goyal2022test,chen2022contrastive}, or some specific parameters of the network, such as batch-normalization \cite{sita, hu2021mixnorm, bn_paper, nado2020evaluating}. Our method falls into the second category where we do not modify the training strategy and do not add any additional head to the network, thus making it applicable to any off-the-shelf text line recognition model. 
\looseness=-1

As these methods are designed for tasks such as classification, it is non-trivial to extend them to sequence learning tasks, which is the focus of this paper. We still adapt some of these methods to the problem at hand, and compare with them in Section \ref{sec:exp}. Most of these methods do not use any additional source of information beyond the network's prediction to improve. In this work, we leverage on a language model to improve beyond the optical model's performance.
\looseness=-1

{\flushleft \textbf{Domain Adaptation for Text Line Recognition.}}
Compared to TTA for classification and segmentation tasks where the notion of style is limited, in text line recognition, the notion of style is quite prevalent, as it can vary widely across writers. There have been some works on adaptation of text line recognition models, which can be classified into three broader categories. First, adapting to an entire dataset with a lot of unlabeled samples \cite{syn2real,tang2022domain,Fogel2020ScrabbleGANSV,seq2seqdomain}. Most of the losses and techniques used in such works are analogous to those used in classification and segmentation tasks in literature. The second category of work \cite{wang2022fast,kohut2023towards} assumes access to writer identification information while training the source model, and trains a separate module to extract writer-specific information. Finally, the third category of work operates in the few shot adaptation technique, assuming access to a few labeled instances for new test writers \cite{metahtr}. 

Contrary to aforementioned works, our method does not need access to the source dataset, no information about writers during training, no labeled data for the test writers, and adapts to only one image of handwriting at test time. This makes our model useful for real-world scenarios and applicable to any off-the-shelf model, without making any changes to the model training. 

\section{Methodology}
We first formally define the problem statement, then explain the architecture of the model we use and then describe the proposed approach of test time adaptation.

\subsection{Problem Statement}
Consider that we are given a text line recognition model, which given an image, predicts the text in it, which is a sequence of  characters, i.e., $f_\theta: \boldsymbol{\mathrm{x}} \rightarrow \boldsymbol{\mathrm{y}}$, where $\boldsymbol{\mathrm{x}} \in \mathbb{R}^{h \times w \times 3}$, and $\boldsymbol{\mathrm{y}} \in \mathbb{V}^n$, where $\mathbb{V}$ is vocabulary set of character, and $n$ is variable depending on the model's prediction. %We resize the image to have the height of size $h=40$, maintaining the aspect ratio. 

We consider the problem of test-time adaptation given an image of a page consisting of multiple text lines. We  can apply off-the-shelf text line detectors to get a list of text lines $\{\boldsymbol{\mathrm{x}}_i\}_{i=1}^m$. We can then apply the text line recognizer $f_\theta$ on the individual lines to get the predicted text. This line recognizer model would generally perform well on handwriting styles that it has seen before, or are similar to such styles. But when encountered with new styles of handwriting, or corruptions on such handwritten images, the model's performance may degrade. Our objective is to adapt the model $f_\theta$ so that it performs better on the test image than the source model. In our setting, after adapting to a test image, we reset the model back to the source model before adapting to a new image. 

\subsection{Model Architecture}
Text line recognition models generally consist of two parts - an encoder, which takes as input a raw image and outputs features or logits, and a decoder, which produces a sequence of output characters from the features generated by the encoder. The decoder can utilize an explicit language model to encode knowledge about the domain to correct the errors made by the optical model. There are two broad categories of models - end-to-end encoder-decoder models \cite{li2021trocr}, and de-coupled encoder decoder models trained separately \cite{Diaz2021RethinkingTL}. In the latter, the optical model is trained on image data using CTC loss, and the language model is trained on only text data. Typically, the two are combined during inference using scores from both the models. In this work, we use this de-coupled model as the source model, as it is often light-weight, and the de-coupled language model decoder allows a plug-and-play approach for new domains. We follow the self-attention based model architecture of \cite{Diaz2021RethinkingTL}. The composite text line recognizer is represented as $f_\theta = h \circ g_\theta$.

{\flushleft \textbf{Optical Encoder} ($g_\theta$)} comprises of a convolutional neural network (CNN) like MobileNet \cite{Howard2017MobileNetsEC}, followed by multiple transformer-like multi-headed self-attention layers \cite{Vaswani2017AttentionIA} to capture longer range context. Finally, a linear classifier is used to predict the symbols for every frame. The input to this block is $\boldsymbol{\mathrm{x}} \in \mathbb{R}^{h \times w \times 3}$, and the output is logits $g_\theta(\boldsymbol{\mathrm{x}}) \in \mathbb{R}^{w' \times C}$, where $w'$ is a downsampled version of $w$, and $C$ is the number of character symbols in the vocabulary. This model is learned using CTC loss \cite{graves2006connectionist} between the ground-truth label string and the logit.

{\flushleft \textbf{Language Model Decoder} ($h$)} takes in the encoder network's logits and combines it with the language model to decode text content from it. The decoded string $\boldsymbol{\mathrm{y}^*}$ can be obtained by optimizing the following - 
\begin{equation}
    \boldsymbol{\mathrm{y}}^* = \argmax_{\boldsymbol{\mathrm{y}}} p(\boldsymbol{\mathrm{y}}|\boldsymbol{\mathrm{x}}) p(\boldsymbol{\mathrm{y}})^\alpha 
    \label{eqn:lm_decoder}
\end{equation}
where $\alpha$ is the weight of the language model. It is interesting to note that the formulation of the CTC algorithm \cite{graves2006connectionist} is such that multiple strings can map to the same final decoded output. An approximate solution to the above problem is obtained by using beam decoding which combines the language model scores, $p(\boldsymbol{\mathrm{y}})$ and optical model scores, $p(\boldsymbol{\mathrm{y}}|\boldsymbol{\mathrm{x}})$ at every step of decoding. 

\subsection{Test-Time Adaptation}
Given an image consisting of multiple lines as input, we want to adapt the model and refine its predictions such that it performs better than when we apply the original source model only. Our adaptation process consists of two parts - updating the optical model, with a computationally efficient language model in the loop (local context), and finally tuning the predictions for every line using a large language model (global context). We next discuss these in detail. 

\subsubsection{Adapting the Optical Model}

As discussed before, there are often idiosyncrasies in an individual's handwriting which are typically consistent. Understanding them would help the model to adapt and personalize the predictions. We develop a self-training mechanism to automatically identify such idiosyncrasies and update the optical model. It is interesting to note that the optical model alone can make certain errors, which the language model corrects. This acts as a feedback signal to the optical model and improves its performance. However, the confidence of the model can vary across lines, and we would want the model to self-improve by progressively learning from confident lines. We next define the measure of confidence we use in our algorithm. 

{\flushleft \textbf{Confidence of the Optical Model.}}
The predictions of deep neural networks have been often found to be incorrect with high confidence \cite{nguyen2015deep}. Because of the highly non-linear nature of neural networks, small perturbations in inputs have been shown to have huge differences in outputs \cite{goodfellow2014explaining}. We hypothesize that smoother transitions in the model's outputs owing to changes in the input, leads to better correlation between confidence and correctness. In other words, if we perturb the image by a small amount, the predictions should not be perturbed significantly. We use this idea and formalize a measure to assign confidence to every line prediction in the image. This can be formally represented as follows,
\begin{equation}
    c(f_\theta(\boldsymbol{\mathrm{x}})) = 1- d(f_\theta(\boldsymbol{\mathrm{x}}), f_\theta(\boldsymbol{\mathrm{\hat{x}}}))
    \label{eqn:conf}
\end{equation}
where $\boldsymbol{\mathrm{\hat{x}}}$ is an augmented version of the image $\boldsymbol{\mathrm{x}}$, and $d()$ can be any distance measure. In our algorithm, we use the normalized edit distance (NED) ($=\nicefrac{\text{EditDist}(f_\theta(\boldsymbol{\mathrm{x}}), f_\theta(\boldsymbol{\mathrm{\hat{x}}}))}{|f_\theta(\boldsymbol{\mathrm{x}})|}$) between the two strings as the distance measure. We use very light augmentations to obtain  $\boldsymbol{\mathrm{\hat{x}}}$, as we want to judge local smoothness of the function. More details are discussed in Section \ref{sec:exp}. 

{\flushleft \textbf{Self-Training Loss.}}
The CTC loss used to train a text line recognizer is at a line level, rather than at a frame level (analogous to pixels in segmentation). Hence, we can only gather self-labels at a line level. Given an image, we first extract the lines from it using any off-the-shelf text line detector \cite{Diaz2021RethinkingTL} to obtain a list of lines $\{\boldsymbol{\mathrm{x}}_i\}_{i=1}^m$. We pass these lines through the encoder and the decoder to obtain self-labels $\{\boldsymbol{\mathrm{\hat{y}}}_i = f_\theta(\boldsymbol{\mathrm{x}})\}_{i=1}^m$. As we only train the optical model using self-training, we can then compute the CTC loss, $\mathcal{L}_{CTC}$ between the self-labels (output of the decoder) and the output of the encoder, i.e., the optical model. Now, as the model may not be equally confident for all lines, we apply the confidence of the model in Eqn \ref{eqn:conf} to weight the losses. The optimization problem we solve can be represented as follows:
\begin{equation}
    \theta^* = \argmin_\theta \sum_{i=1}^m c(f_\theta(\boldsymbol{\mathrm{x}}_i))\mathcal{L}_{CTC}\big(g_\theta(\boldsymbol{\mathrm{x}}_i), \boldsymbol{\mathrm{\hat{y}}}_i\big)
    \label{eqn:selftrain}
\end{equation}

Note that all learnable parameters of the network pertain to $g_\theta$, and the above equation optimizes it only. One can also compute the self-labels using just $g_\theta$, i.e., from the optical model's output itself instead of using the language model incorporated decoder $h \circ g_\theta$. But, we hypothesize that the language model acts as an additional supervisory signal, which helps to improve the performance. We also show that ablation in Section \ref{sec:exp}.
We do not compute any gradients through the confidence function $c()$, but they are updated in every step through forward propagation, as is discussed next. 

{\flushleft \textbf{Progressive Updates.}}
Although we use a confidence measure $c()$ to weight the CTC loss for every line, the confidence metric as well as the self-labels gets better as we adapt. Hence, we progressively update the model starting with high confidence predictions, while updating the self-labels as well as the confidence measure in each iteration. This would allow the model to identify the writer's style and self-improve, compared to fixing the self-labels and the confidence measure once. We start from the most confident lines, and in each iteration, we keep adding the next most confident ones progressively. Thus, considering we back-propagate for $K$ iterations in total, in the iteration number $k \leq K$, we progressively add $m_k = (\nicefrac{k}{K})m$ samples, which have the highest values of the confidence metric, to backpropagate and optimize Eqn \ref{eqn:selftrain}. 

Our method to adapt the optical model is shown in Algorithm \ref{algo:optical}. In each progressive iteration, we first make a forward pass to compute the self-labels $\boldsymbol{\mathrm{\hat{y}}}_i$, and the confidence measure $c(f_\theta(\boldsymbol{\mathrm{x}}_i))$, and then backpropagate only using the fraction of the samples based on the confidence measure. Note that our model learns from self-labels throughout all iterations, and there is a possibility of divergence if the initial predictions turn out to be more incorrect than not. Hence, after adaptation, we compare the final prediction with the initial prediction and if the edit distance between the two is greater than a certain threshold (=$0.75$ used in all experiments), then we use the initial prediction itself. 

The above process only updates the encoder, i.e., the optical model. We next discuss how we can tune the predictions using a large language model, instead of the computationally cheap LM used in the decoder, $h$ here. 

\begin{algorithm}
\small{
    \caption{Adapting the Optical Model}
    \label{algo:optical}
    \textbf{Input:} Source model: $f_\theta = h \circ g_\theta$, List of lines: $\{\boldsymbol{\mathrm{x}}_i\}_{i=1}^m$ \\
    \textbf{Output:} Updated model: $f_\theta = h \circ g_{\theta^*}$ 
    \begin{algorithmic}[1]
    \State $\theta_0 \leftarrow \theta$ \smallskip
    \For {\texttt{$k = 1 \dots K$}} \smallskip
    \State $\boldsymbol{\mathrm{\hat{x}}}_i \leftarrow \text{Augment} (\boldsymbol{\mathrm{x}}_i), \forall i \in [1, m]$ \smallskip
    \State {\em $//$ Forward Propagate} \smallskip
    \State Obtain $f_\theta(\boldsymbol{\mathrm{x}}_i)$ and $f_\theta(\boldsymbol{\mathrm{\hat{x}}}_i)$ $\forall i \in [1, m]$ \smallskip
    \State $c(f_\theta(\boldsymbol{\mathrm{x}}_i)) \leftarrow 1- d(f_\theta(\boldsymbol{\mathrm{x}}_i), f_\theta(\boldsymbol{\mathrm{\hat{x}}}_i)) , \forall i \in [1, m]$ \smallskip
    \State {\em $//$ Back Propagate}
    \State \vspace*{-6mm}
    \begin{align}
        \theta_k & \leftarrow \theta_{k-1} -\nonumber \\
        & \eta \sum_{i \in \text{top-m$_k$}} c(f_\theta(\boldsymbol{\mathrm{x}}_i)) \nabla_\theta \mathcal{L}_{CTC}\big(g_\theta(\boldsymbol{\mathrm{x}}_i), \boldsymbol{\mathrm{\hat{y}}}_i\big)
    \end{align}
    % \State $\theta_k \leftarrow \theta_{k-1} - \eta \sum_{i \in \text{top} } c(f_\theta(\boldsymbol{\mathrm{x}}_i)) \nabla_\theta \mathcal{L}_{CTC}\big(g_\theta(\boldsymbol{\mathrm{x}}_i), \boldsymbol{\mathrm{\hat{y}}}_i\big)$
    \vspace*{-4mm}
    \EndFor
    \State $\theta^* \leftarrow \theta_K$
    \end{algorithmic}
}
\end{algorithm}

\subsubsection{Adaptation using a Large Language Model}
The optical model looks at local context to make a prediction. However, longer length context often helps to correct some of the errors. Large Language Models (LLM) using words tokens instead of characters offers a good mechanism to extract longer context information present in natural text. While an LLM can be used in Eqn \ref{eqn:lm_decoder} itself, that can be computationally expensive. To avoid that, we extract top-k predictions from the CTC decoder $\{\boldsymbol{\mathrm{y}^i}\}_{i=1}^k$, and then re-score the lines using an LLM. We use a pre-trained FlanT5-XL model \cite{chung2022scaling} as the language model for re-scoring. The log-likelihood score of the LLM can be computed as follows:
\begin{equation}
    \mathcal{L}_{LLM} =  -\sum_{j=1}^W \log P( \boldsymbol{\mathrm{w}}_j | \boldsymbol{\mathrm{w}}_1, ... ,\boldsymbol{\mathrm{w}}_{j-1}) 
    \label{eqn:llm_loss}
\end{equation}

\noindent
where $\boldsymbol{\mathrm{w}}_i$ represents the $j^{th}$ word token in a sentence $\boldsymbol{\mathrm{y}}$. Using only the LLM loss may make it hallucinate, and thus to remain grounded to the actual text in the image, we add the optical score $\mathcal{L}_{opt}$, which is a combination of the scores from the optical logits and the n-gram LM used in the decoder. Finally, we take a weighted sum of these two scores to pick the best candidate amongst the generated top-k predictions as follows - 
\begin{equation}
    \mathrm{best\_candidate} = \argmin_{i} \mathcal{L}_{opt}^i + w*\mathcal{L}_{LLM}^i
    \label{eqn:llm_weighted_sum}
\end{equation}
where $w$ is the weight given to the LLM score, which is a hyper parameter (set to $0.5$ for all our experiments).

\section{Experiments} \label{sec:exp}
\begin{figure}
     \centering
     \begin{subfigure}[b]{0.25\columnwidth}
         \centering
         \includegraphics[height=2.5cm]{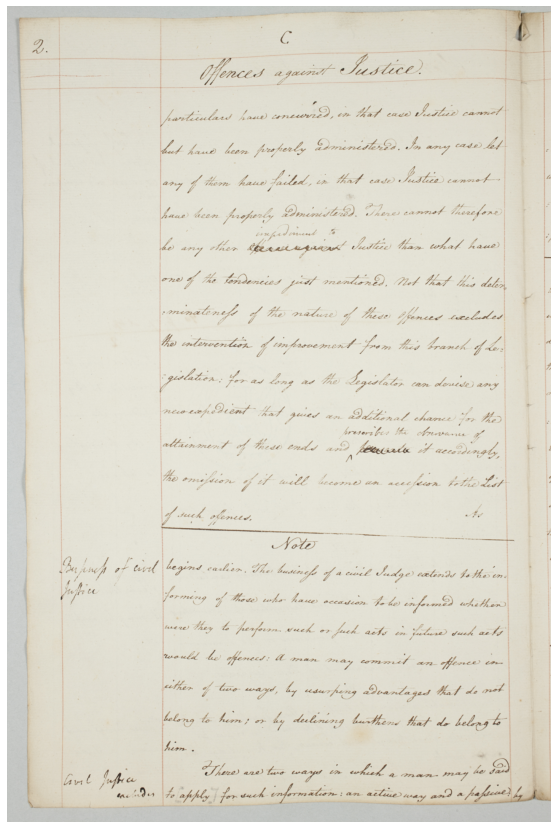}
         \caption{\small ICDAR2015}
         \label{fig:ICDAR2015-HTR}
     \end{subfigure}
     \hfill
     \begin{subfigure}[b]{0.25\columnwidth}
         \centering
         \includegraphics[height=2.5cm]{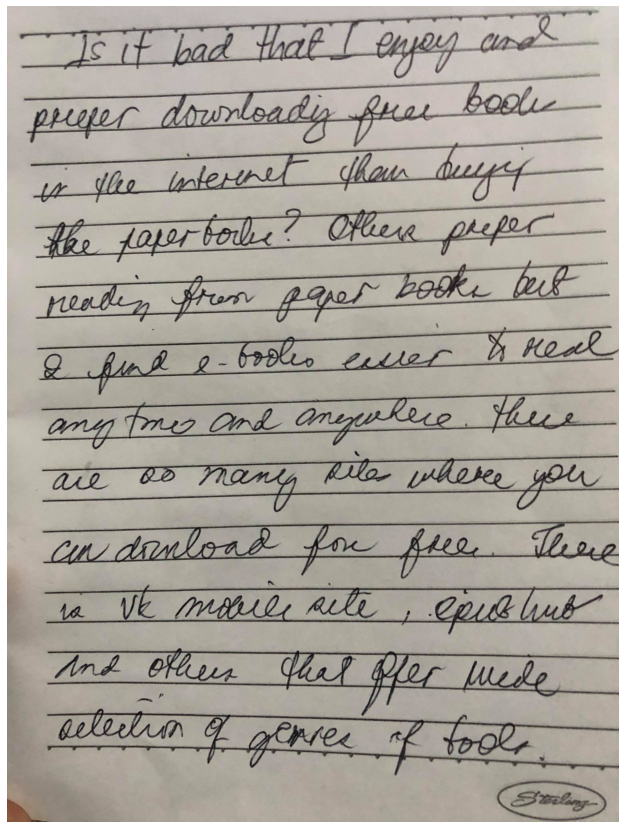}
         \caption{\small GNHK}
         \label{fig:GNHK}
     \end{subfigure}
     \hfill
     \begin{subfigure}[b]{0.25\columnwidth}
         \centering
         \includegraphics[height=2.5cm]{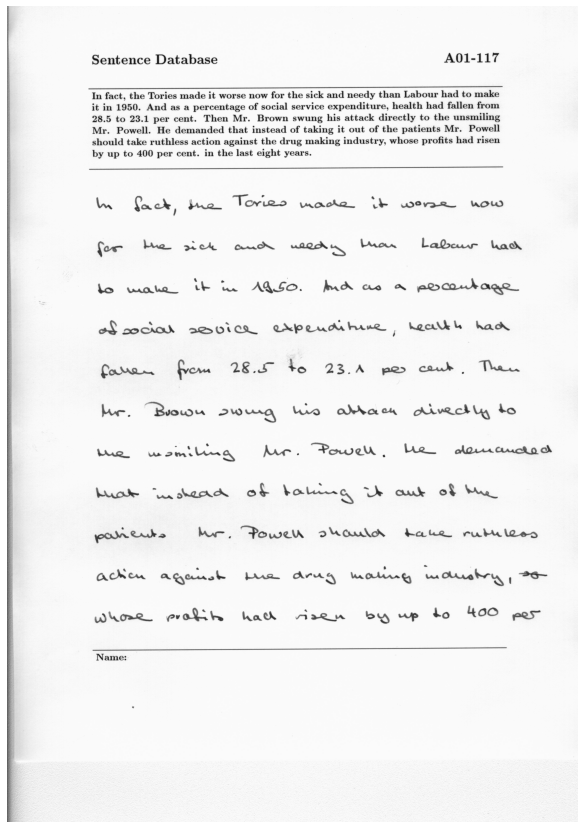}
         \caption{\small IAM}
         \label{fig:IAM}
     \end{subfigure}
     \begin{subfigure}[b]{0.3\columnwidth}
         \centering
         \includegraphics[height=3.2cm]{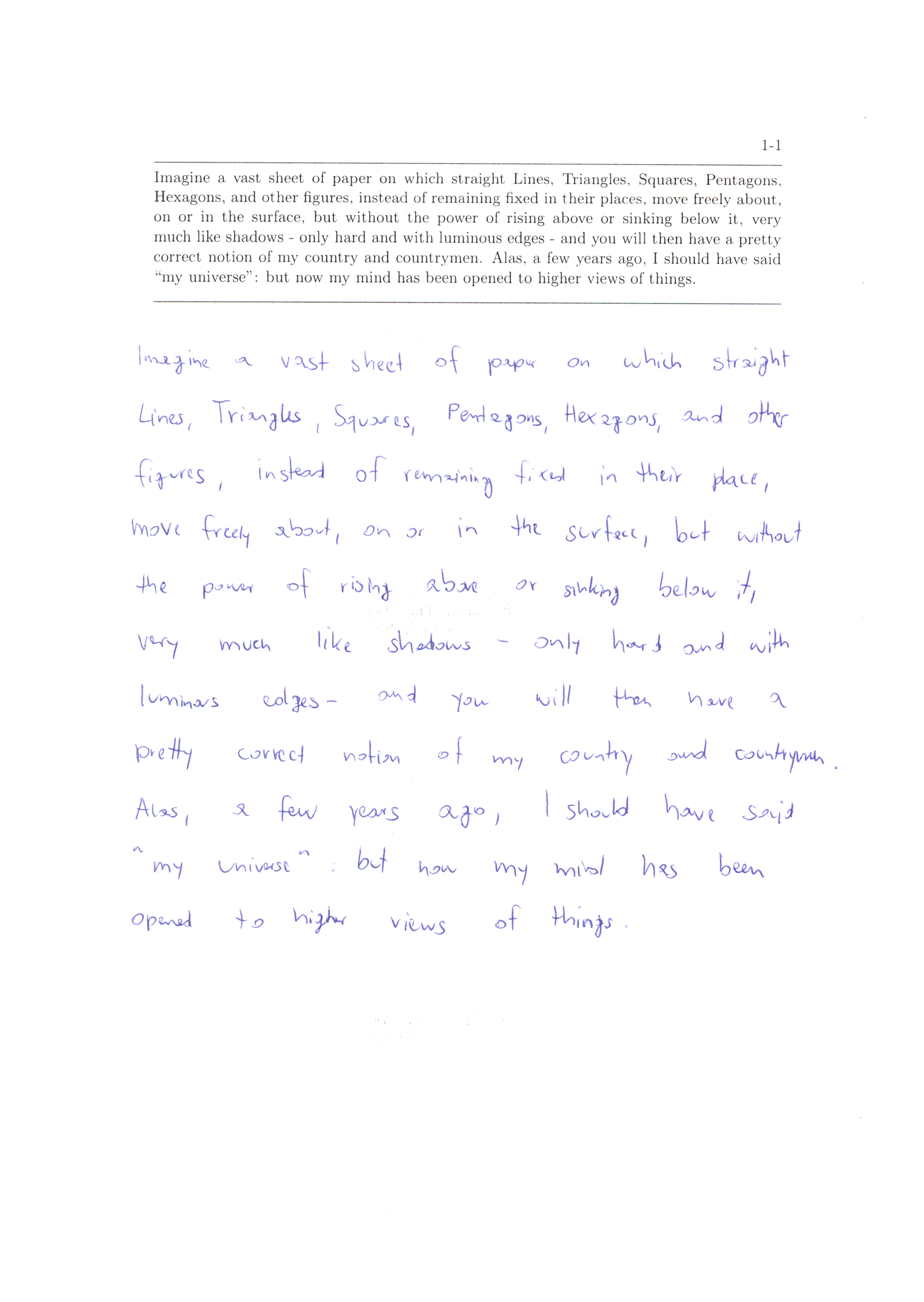}
         \caption{\small CVL}
         \label{fig:CVL}
     \end{subfigure}
     \begin{subfigure}[b]{0.3\columnwidth}
         \centering
         \includegraphics[height=3.2cm]{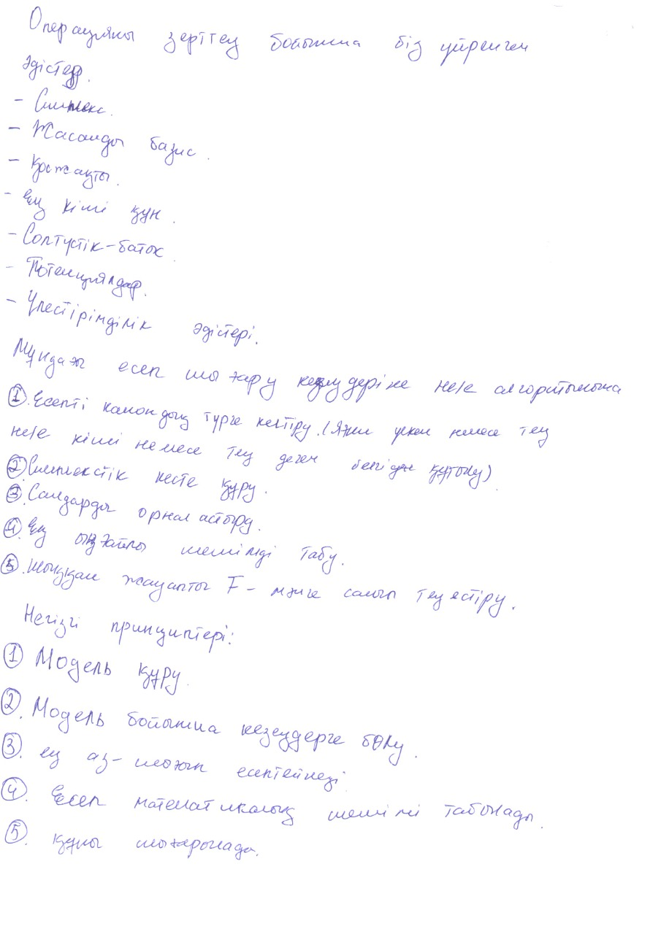}
         \caption{\small KOHTD}
         \label{fig:KOHTD}
     \end{subfigure}
    \caption{Exemplars from the datasets used in this paper.}
    \label{fig:dataset}
\end{figure}
\newcommand\lineheight{0.33cm}
\begin{figure*}
     \centering
     \begin{subfigure}[b]{0.19\textwidth}
         \centering
         \includegraphics[height=\lineheight]{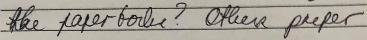}
         \vspace{-2mm}
         \caption{\small Original}
     \end{subfigure}
     \begin{subfigure}[b]{0.19\textwidth}
         \centering
         \includegraphics[height=\lineheight]{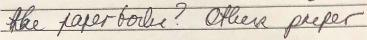}
         \vspace{-2mm}
         \caption{\small Brightness}
     \end{subfigure}
     \begin{subfigure}[b]{0.19\textwidth}
         \centering
         \includegraphics[height=\lineheight]{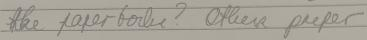}
         \vspace{-2mm}
         \caption{\small Contrast}
     \end{subfigure}
     \begin{subfigure}[b]{0.19\textwidth}
         \centering
         \includegraphics[height=\lineheight]{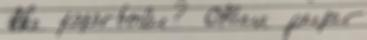}
         \vspace{-2mm}
         \caption{\small Defocus Blur}
     \end{subfigure}
     \begin{subfigure}[b]{0.19\textwidth}
         \centering
         \includegraphics[height=\lineheight]{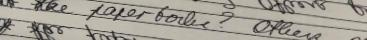}
         \vspace{-2mm}
         \caption{\small Elastic}
     \end{subfigure}

     \begin{subfigure}[b]{0.19\textwidth}
         \centering
         \includegraphics[height=\lineheight]{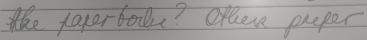}
         \vspace{-2mm}
         \caption{\small Fog}
     \end{subfigure}
     \begin{subfigure}[b]{0.19\textwidth}
         \centering
         \includegraphics[height=\lineheight]{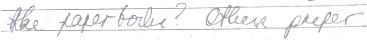}
         \vspace{-2mm}
         \caption{\small Frost}
     \end{subfigure}
     \begin{subfigure}[b]{0.19\textwidth}
         \centering
         \includegraphics[height=\lineheight]{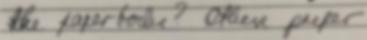}
         \vspace{-2mm}
         \caption{\small Gaussian Blur}
     \end{subfigure}
     \begin{subfigure}[b]{0.19\textwidth}
         \centering
         \includegraphics[height=\lineheight]{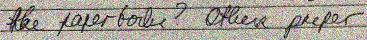}
         \vspace{-2mm}
         \caption{\small Gaussian Noise}
     \end{subfigure}
     \begin{subfigure}[b]{0.19\textwidth}
         \centering
         \includegraphics[height=\lineheight]{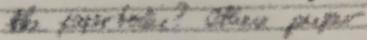}
         \vspace{-2mm}
         \caption{\small Glass Blur}
     \end{subfigure}
     
     \begin{subfigure}[b]{0.19\textwidth}
         \centering
         \includegraphics[height=\lineheight]{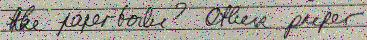}
         \vspace{-2mm}
         \caption{\small Impulse Noise}
     \end{subfigure}
     \begin{subfigure}[b]{0.19\textwidth}
         \centering
         \includegraphics[height=\lineheight]{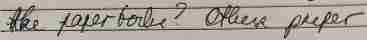}
         \vspace{-2mm}
         \caption{\small Jpeg}
     \end{subfigure}
     \begin{subfigure}[b]{0.19\textwidth}
         \centering
         \includegraphics[height=\lineheight]{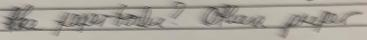}
         \vspace{-2mm}
         \caption{\small Motion Blur}
     \end{subfigure}
     \begin{subfigure}[b]{0.19\textwidth}
         \centering
         \includegraphics[height=\lineheight]{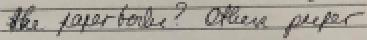}
         \vspace{-2mm}
         \caption{\small Pixelate}
     \end{subfigure}
     \begin{subfigure}[b]{0.19\textwidth}
         \centering
         \includegraphics[height=\lineheight]{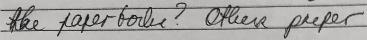}
         \vspace{-2mm}
         \caption{\small Saturate}
     \end{subfigure}
     
     \begin{subfigure}[b]{0.19\textwidth}
         \centering
         \includegraphics[height=\lineheight]{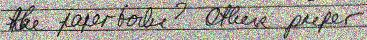}
         \vspace{-2mm}
         \caption{\small Shot Noise}
     \end{subfigure}
     \begin{subfigure}[b]{0.19\textwidth}
         \centering
         \includegraphics[height=\lineheight]{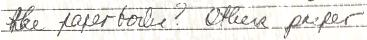}
         \vspace{-2mm}
         \caption{\small Snow}
     \end{subfigure}
     \begin{subfigure}[b]{0.19\textwidth}
         \centering
         \includegraphics[height=\lineheight]{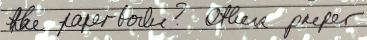}
         \vspace{-2mm}
         \caption{\small Spatter}
     \end{subfigure}
     \begin{subfigure}[b]{0.19\textwidth}
         \centering
         \includegraphics[height=\lineheight]{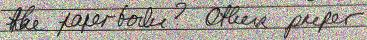}
         \vspace{-2mm}
         \caption{\small Speckle Noise}
     \end{subfigure}
     \begin{subfigure}[b]{0.19\textwidth}
         \centering
         \includegraphics[height=\lineheight]{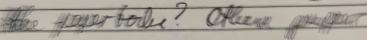}
         \vspace{-2mm}
         \caption{\small Zoom Blur}
     \end{subfigure}
    \caption{An example original image line and $19$ corruptions used from \cite{hendrycks2019benchmarking}. %These corruptions are applied to all images on all datasets.
    }
    \label{fig:lines}
    \vspace{-1 em}
\end{figure*}
To showcase the effectiveness of our method, we perform thorough experimentation and qualitative analysis on five benchmark datasets.

{\flushleft \textbf{Datasets}}: 

{\flushleft \contour{black}{ICDAR2015-HTR} \cite{icdar2015}} consists of handwritten historical documents in English. This is a particularly challenging dataset due to the variance in writing styles and quality of images. We use the entire dataset of $433$ images (train, test, and validation) for testing. We use the split of the dataset which contains line-level annotations.

{\flushleft \contour{black}{GoodNotes Handwriting Kollection (GNHK)} \cite{goodnotes}} is a handwritten English text dataset, sourced from different regions in the world. It contains various types of texts, like diary notes, shopping lists, etc. captured through a mobile phone. This dataset has only word-level annotations, with line ids. We convert that to line annotation by concatenating the word annotations using spaces, and then joining the word images. We use the train and test split consisting of $687$ images for evaluation. We omit lines which are printed and have math symbols, as there are no transcriptions for the latter. 
{\flushleft \contour{black}{IAM Handwriting Database} \cite{iam}} consists of English handwritten text consisting of $1539$ pages from $657$ unique writers. The images for this dataset are much clearer than the other datasets and also has line-level annotations. We use the entire dataset for evaluation. 

{\flushleft \contour{black}{CVL Database} \cite{kleber2013cvl}} consists of English handwritten text consisting of $1598$ pages from $7$ handwritten texts (1 German and 6 English Texts). $310$ writers participated in the dataset. 27 of which wrote 7 texts and 283 writers had to write 5 texts. We use lines from both train and test splits to evaluate the efficacy of our approach. 

{\flushleft \contour{black}{KOHTD} \cite{kohtd}} consists of a large collection of exam papers filled by students in the Kazakh Language (99\%) and Russian Language (1\%). It contains a total of 1891 images, containing word level annotations for each page. A sample image from each of the datasets is shown in Figure~\ref{fig:dataset}.

\hspace{0.7cm}

{\flushleft \textbf{Corruptions}}:
In real world scenarios, image acquisition generally adds corruptions to the image such as blur, noise, compression, etc. Following \cite{hendrycks2019benchmarking}, we create $19$ different corrupted versions for each of the five datasets (Figure~\ref{fig:lines}). 
%We report the results on both the original and the corrupted versions. 
\looseness=-1

{\flushleft \textbf{Implementation Details}}:
The source model is trained in the same way and on the same datasets as in \cite{Diaz2021RethinkingTL}. This model is strong enough as it is trained on lots of labeled and synthetically generated text lines. To calculate the confidence measure for all lines, we augment them using three augmentations, viz, mean filter, median filter and sharpness. Our n-gram model is also similar to \cite{Diaz2021RethinkingTL} ($n=9$). 
For test-time adaptation, we use SGD, with a momentum of 0.9, and a learning rate of $10^{-3}$. All lines in an image form a single batch of data, to which the model is to be adapted. We follow the same model architecture and configurations as in \cite{Diaz2021RethinkingTL}. Only the optical encoder $g_\theta$ is adapted to a writer's handwriting.

{\flushleft \textbf{Baselines}}: To the best of our knowledge, this is the first work on test-time adaptation of text line recognition models from only a single test image, without accessing any source data to adapt or without learning any writer identification models while training the source model. As there is no direct baseline in the literature for this use case, we consider three strong baselines from the TTA literature for classification and segmentation, namely BatchNormalization Adaptation (BN) \cite{bn_paper}, TENT \cite{wang2021tent}, and Prediction Time Normalization (PTN) \cite{nado2020evaluating}. These are suited for our problem statement as they do not need any additional changes to the source model training pipeline. While BN and PTN are trivial to extend to this use case, the same is not the case for TENT, which was designed for classification tasks. TENT minimizes the entropy of the predictions by modifying only the affine parameters of the batch normalization layer. As the problem at hand involves a many-to-one mapping from input frames to final sequence, computing the entropy over unique mappings is non-trivial and computationally intractable. Thus, we minimize the mean entropy of all the frames over all lines in the page. We do it for the same number of iterations as in our algorithm.
\looseness=-1

\begin{table}[t]
            \small
			\caption{\small Performance (CER - lower better) comparison of the proposed method with baselines. ``Original" denotes the performance on the non-corrupted dataset and ``Corrupted" denotes the average performance over all corruptions. 
			The proposed approach outperforms all, both on original and corrupted versions.
			%The performance improvement of our method is not only on corrupted datasets, which is more challenging, but also on the original dataset where the distribution is closer to the train set.
			}
			\label{table:seg_iter}
			\centering
			\renewcommand{\arraystretch}{1.2}
	        \setlength{\tabcolsep}{4pt}
	        %\resizebox{0.6\textwidth}{!}{
			\begin{tabular}{l|cccccc}
		    \toprule
		    Dataset & & Source & BN & PTN & TENT & Ours \\
		    \midrule
		    \multirow{2}{*}{IC15-HTR}  & Original & 14.4 & 17.0 & 46.8 & 14.3 & \textbf{13.2}\\
		     & Corrupted & 27.6 & 30.6 & 61.6 & 27.6 & \textbf{25.3}\\
		    \midrule
		    \multirow{2}{*}{GNHK} & Original & 14.0 & 14.2 & 21.2 & 14.3 & \textbf{13.3}\\
		     & Corrupted & 24.4 & 24.2 & 38.6 & 25.5 & \textbf{23.0}\\
		    \midrule
		    \multirow{2}{*}{IAM} & Original & 6.0 & 6.1 & 12.5 & 6.1 & \textbf{5.4}\\
		     & Corrupted & 15.4 & 15.1 & 28.7 & 15.5 & \textbf{13.8}\\
		     \midrule
		    \multirow{2}{*}{CVL} & Original & 8.6 & 8.7 & 14.6 & 8.6 & \textbf{7.7}\\
		     & Corrupted & 26.4 & \textbf{25.8} & 41.8 & 26.4 & 25.9\\
		      \midrule
		    \multirow{2}{*}{KOHTD} & Original & 23.3 & 23.3 & 29.4 & 23.3 & \textbf{16.8}\\
		     & Corrupted & 32.0 & 31.9 & 40.8 & 32.0 & \textbf{25.2}\\
		   \bottomrule
	    \end{tabular}
	    %}
\end{table}

\begin{figure}
     \centering
    %  \begin{subfigure}[b]{\columnwidth}
    %      \centering
    %      \includegraphics[scale=0.45]{./figures/icdar_deg.pdf}
    %      \caption{\small ICDAR2015}
    %      \label{fig:icdar-deg}
    %  \end{subfigure}
     \begin{subfigure}[b]{\columnwidth}
         \centering
         \includegraphics[scale=0.5]{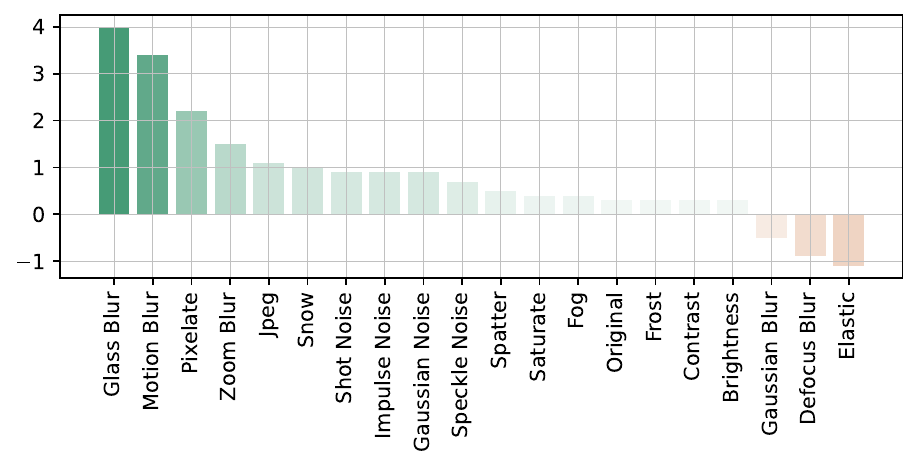}
         \caption{\small GNHK}
         \label{fig:gnhk-deg}
     \end{subfigure}
     \begin{subfigure}[b]{\columnwidth}
         \centering
         \includegraphics[scale=0.45]{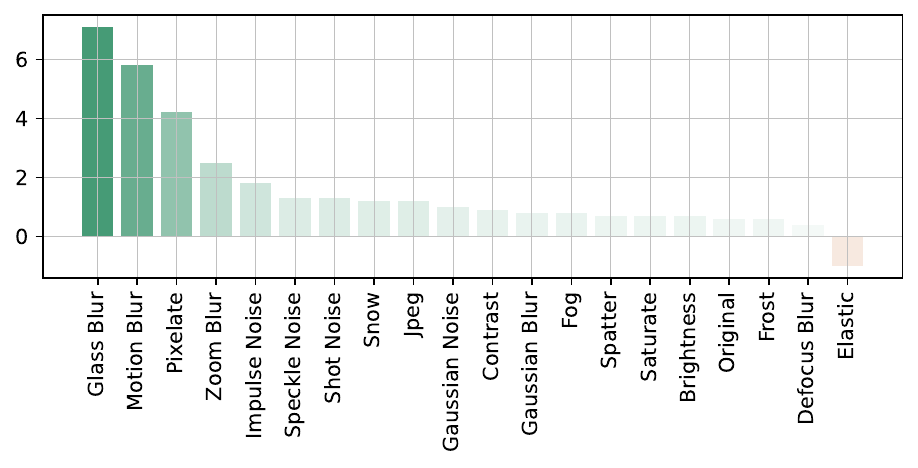}
         \caption{\small IAM}
         \label{fig:iam-deg}
     \end{subfigure}
    %  \begin{subfigure}[b]{\columnwidth}
    %      \centering
    %      \includegraphics[scale=0.5]{./figures/cvl_deg.pdf}
    %      \caption{\small CVL}
    %      \label{fig:cvl-deg}
    %  \end{subfigure}
     \begin{subfigure}[b]{\columnwidth}
         \centering
         \includegraphics[scale=0.5]{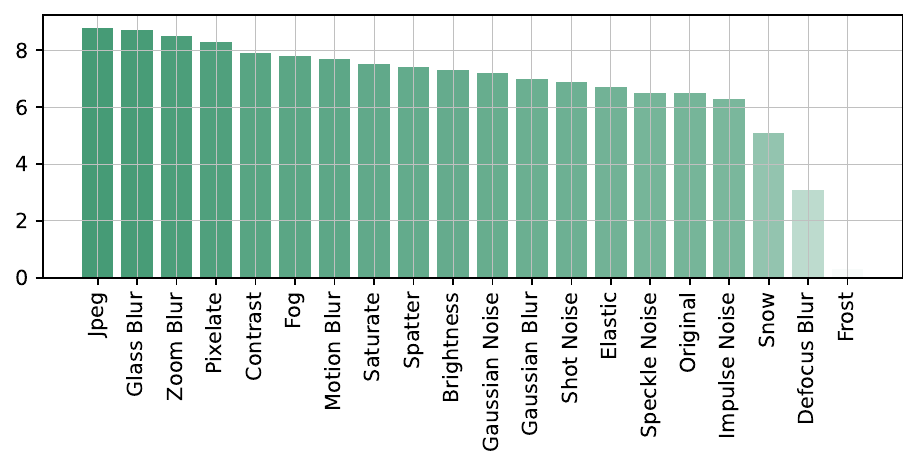}
         \caption{\small KOHTD}
         \label{fig:kohtd-deg}
     \end{subfigure}
        \caption{Absolute improvement (in $\%$) obtained with our approach for various corruptions. Apart from minor regressions for a couple of corruption types, the proposed approach shows significant improvement across the board.}
        \label{fig:degradation}
\end{figure}

{\flushleft \textbf{Comparison with baselines:}}
% \subsection{Results and Analysis}
Table~\ref{table:seg_iter} shows the performance of our approach compared against the three baselines and the source model. For all the five datasets, the proposed approach outperforms all others. The quantum of improvement is further established when the experiment is repeated on the corrupted versions of the datasets, as shown in Figure~\ref{fig:lines}. The relative improvements across the $19$ corruptions for two Latin and one non-Latin datasets are shown in Figure~\ref{fig:degradation}. Please refer to the appendix for corruption wise improvements. Please note that the original algorithm for TENT uses a batch of more than 100 images, and that too with online updates, i.e., without resetting the model back to the source model after every update, which is what we do in this work. Moreover, we observe that changing the BatchNorm parameters of the network has a significant impact on the performance. We can see this from the performance of BN which combines the target statistics (mean and variance) with the source statistics using a convex combination. PTN takes this to the extreme by completely replacing the source statistics with the target's.

\begin{table}[t]
    \small
	\caption{\small Ablation for self-training algorithm to update the optical model. All methods are trained for the same number of iterations. ``NED" here denotes the normalized edit distance, which is the distance function used to compute the confidence measure. All performances are average over $20$ datasets (original + $19$ corruptions).}
	\label{table:selftrain_ablation}
	\centering
	\renewcommand{\arraystretch}{1.2}
	\setlength{\tabcolsep}{4pt}
	\resizebox{0.47\textwidth}{!}{
	\begin{tabular}{cccccc}
    \toprule
    Update & \multirow{2}{*}{Progressive} & \multirow{2}{*}{Weighting} & \multicolumn{2}{c}{Confidence} & \multirow{2}{*}{Perf (CER$\downarrow$)} \\
    self-labels & & & CTC & NED & \\
    \midrule
    & & & & & 26.9 \\
    \checkmark & & & & & 26.6 \\
    \checkmark & \checkmark & & \checkmark & & 27.4 \\
    \checkmark & \checkmark & \checkmark & \checkmark & & 25.5 \\
    \checkmark & \checkmark & & & \checkmark & 25.3 \\
    \checkmark & \checkmark & \checkmark & & \checkmark & \textbf{24.7} \\
    \bottomrule
	\end{tabular}
	}
\end{table}

{\flushleft \textbf{Ablation of different choices in self-training}}:
There are several design choices in the proposed self-training algorithm, namely, a) whether to update the self-labels in every learning iteration, b) whether to include the self-labels in the training set progressively instead of taking all of the lines at once, c) whether to weight the loss using a confidence metric, and finally d) the confidence measure we use to choose the lines for progressive updates. We conduct an ablation over these choices to observe their individual importance. Table \ref{table:selftrain_ablation} shows this analysis for ICDAR2015-HTR (please refer to appendix for the other datasets). The augmentation based NED metric we use in our algorithm performs better ($0.8\%$) than just using the CTC loss itself as the confidence metric. Moreover, progressive updates lead to $\textbf{1.6}\%$ improvement than using all of the lines in back-propagation for all iterations. 
\looseness=-1

\begin{table*}[t]
            \small
			\caption{\small Most frequent character replacements made by our TTA algorithm that lead to a better performance.}
			\label{table:changes}
			\centering
			\renewcommand{\arraystretch}{1.2}
	        \setlength{\tabcolsep}{4pt}
	       % \resizebox{0.9\textwidth}{!}{
			\begin{tabular}{l|cccccccccccccccccccc}
		    \toprule
		    Before & {\em o} & {\em l} & {\em a} & {\em a} & {\em n} & {\em e} & {\em i} & {\em m} & {\em e} & {\em e} & {\em T} & {\em a} & {\em n} & {\em u} & {\em a} & {\em s} & {\em o} & {\em d} & {\em e} & {\em r}\\
		    
		    After & {\em e} & {\em t} & {\em o} & {\em e} & {\em r} & {\em o} & {\em e} & {\em n} & {\em i} & {\em a} & {\em t} & {\em u} & {\em e} & {\em e} & {\em r} & {\em e} & {\em a} & {\em t} & {\em s} & {\em n}\\ 
		    \midrule
		    Count & 182 & 155 & 151 & 150 & 149 & 135 & 122 & 108 & 103 & 93 & 92 & 91 & 87 & 87 & 84 & 82 & 80 & 80 & 80 & 78\\
		   \bottomrule
	    \end{tabular}
	    %}
\end{table*}

\begin{table}[t]
    \small
	\caption{\small Importance of updating the optical model, and/or re-scoring the final outputs using an LLM. All performances are average over 20 datasets (original + $19$ corruptions).}
	\label{table:llm_ablation}
	\centering
	\renewcommand{\arraystretch}{1.2}
	\setlength{\tabcolsep}{4pt}
	% \resizebox{0.49\textwidth}{!}{
	\begin{tabular}{ccc}
    \toprule
    Optical Updates & LLM Rescoring & Perf (CER$\downarrow$) \\
    \midrule
    & &  26.9\\
    & \checkmark & 26.5 \\
    \checkmark & & 24.9\\
    \checkmark & \checkmark & \textbf{24.7}\\
    \bottomrule
	\end{tabular}
	% }
\end{table}

{\flushleft \textbf{Ablation of Optical and LLM updates}}:
Our test time adaptation algorithm consists of two parts - adapting the optical model and re-scoring using a large language model. We show an ablation by switching on/off these two blocks in Table \ref{table:llm_ablation}. The first row represents the source model. As we can see, the optical model updates and LLM re-scoring on their own bring about $2.0\%$ and $0.4\%$ improvement on average over all corruptions, and the composite model outperforms the source model by about $\textbf{2.2}\%$.

\begin{figure}
    \centering
    \includegraphics[scale=0.4]{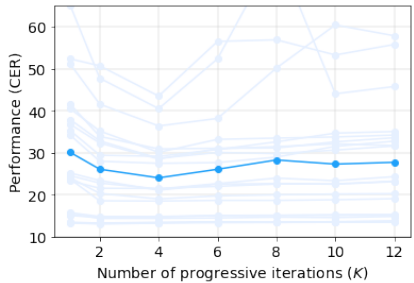}
    \caption{\small \textbf{Performance variation for different number of updates.} Each line shows the performance for one corruption. The average performance is highlighted in the darker shade.}
    \label{fig:num_iter}
    \vspace{-1em}
\end{figure}

{\flushleft \textbf{Ablation of number of iterations}}:
In our approach, if a page has $m$ lines, we progressively add $\nicefrac{m}{K}$ lines based on the confidence metric rank in each iteration, where $K$ is the total number of progressive update iterations. In all our experiments over all datasets, we use $K=4$. Figure~\ref{fig:num_iter} shows the variations in performance for different values of $K$. 
%The mean performance is shown in darker shade and the all other lines are for all the corruptions. 
The performance variations are more apparent for corruptions where the performance is low. If too few updates are used, then there is less scope for improvement through self-labeling. On the other hand, if too many rounds of updates are performed, model outputs tend to  diverge, effectively over-fitting to inaccurate self-labeled data. 

{\flushleft \textbf{What are the changes that the model makes?}}:
Moreover, we analyze the changes in character predictions, the model makes that leads to improvement in performance. We create a list of all replacement edits between the ground-truth and the source model predictions, as well as between the ground-truth and the TTA adapted model. 
We then look at the replacements which are not there in the second list, but there in the first list. 
%This gives us the replacements that the model made, which in turn indicates improvement in performance. 
We analyze this on the ICDAR2015-HTR dataset and present the most frequent replacements in Table~\ref{table:changes}. Most replacements occur among similar looking characters and our TTA algorithm is able to figure them out on the fly. 
\looseness=-1

\begin{figure}
    \centering
    \includegraphics[scale=0.4]{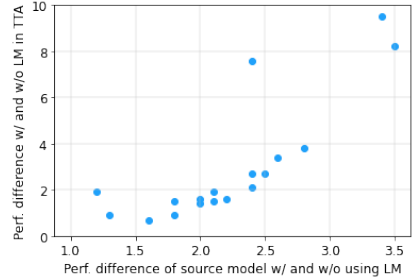}
    \caption{\small \textbf{The role of LM in adapting the optical model.} Each point here denotes the relative improvement the LM introduces for various corrupted versions of the GNHK dataset. The x value of each point is the CER improvement when decoding is done using the LM Decoder vs GreedyDecoder when evaluating the source model. The y-axis shows the CER improvement when our proposed method is used with the LM Decoder vs GreedyDecoder.}
    % Each point denote results for a corrupted dataset of ICDAR2015-HTR. The x-axis of each point is the performance improvement of the source model when LM in CTC decoder is used vs greedy decoder. The y-axis of that point show the performance improvement when LM is also used in TTA vs using greedy decoder in TTA. }

    \label{fig:role_LM}
    \vspace{-1 em}
\end{figure}

{\flushleft \textbf{The role of LM in adapting the optical model}}: As shown in Figure \ref{fig:framework}, our TTA algorithm includes the language model in the loop to update the optical model. This plays an interesting role, compared to other works in TTA which do not use any extra information. As the model is updated using self-training, it can often diverge because of incorrect self-labels. However, the LM acts as a correction module, preventing accidental divergence. We performed an experiment without LM in our TTA algorithm and observed a CER of $29.5\%$ averaged over all corruptions on GNHK, compared to $23.9\%$ when LM is used in TTA. We also plot the performance improvement of the source model with and without using LM in the decoder vs the performance difference with and without using LM in TTA (Figure~\ref{fig:role_LM}). 
% Note that for the after TTA numbers, we use LM to make the final prediction, even though LM is not used in TTA. 
It shows that for corruptions where using LM in decoding offers higher improvement for source model's inference, using LM in TTA shows even greater improvement compared to TTA without LM. 
This is interesting and further highlights the importance of using LM in TTA particularly when the optical model is more confused.
%This is because the self-labels and hence the performance diverges when LM is not used in TTA. 
\looseness=-1

\section{Conclusion}
We introduce the problem of adapting text line recognition models to specific writers using a single test image, without accessing any labeled source data. We develop a method that first progressively adapts the optical model, using an augmentation based confidence function (local context). We further tune the predictions using an LLM which looks at the context of the entire line. Through rigorous experiments and ablation studies on five benchmark datasets, we establish the efficacy of our method.

\newpage
\bibliography{aaai24}

%---------------------------------------------------------------------------------------
\clearpage
\appendix

\section*{\centering\huge Appendix }

\begin{figure*}[h]
    \centering
    \includegraphics[scale=0.5]{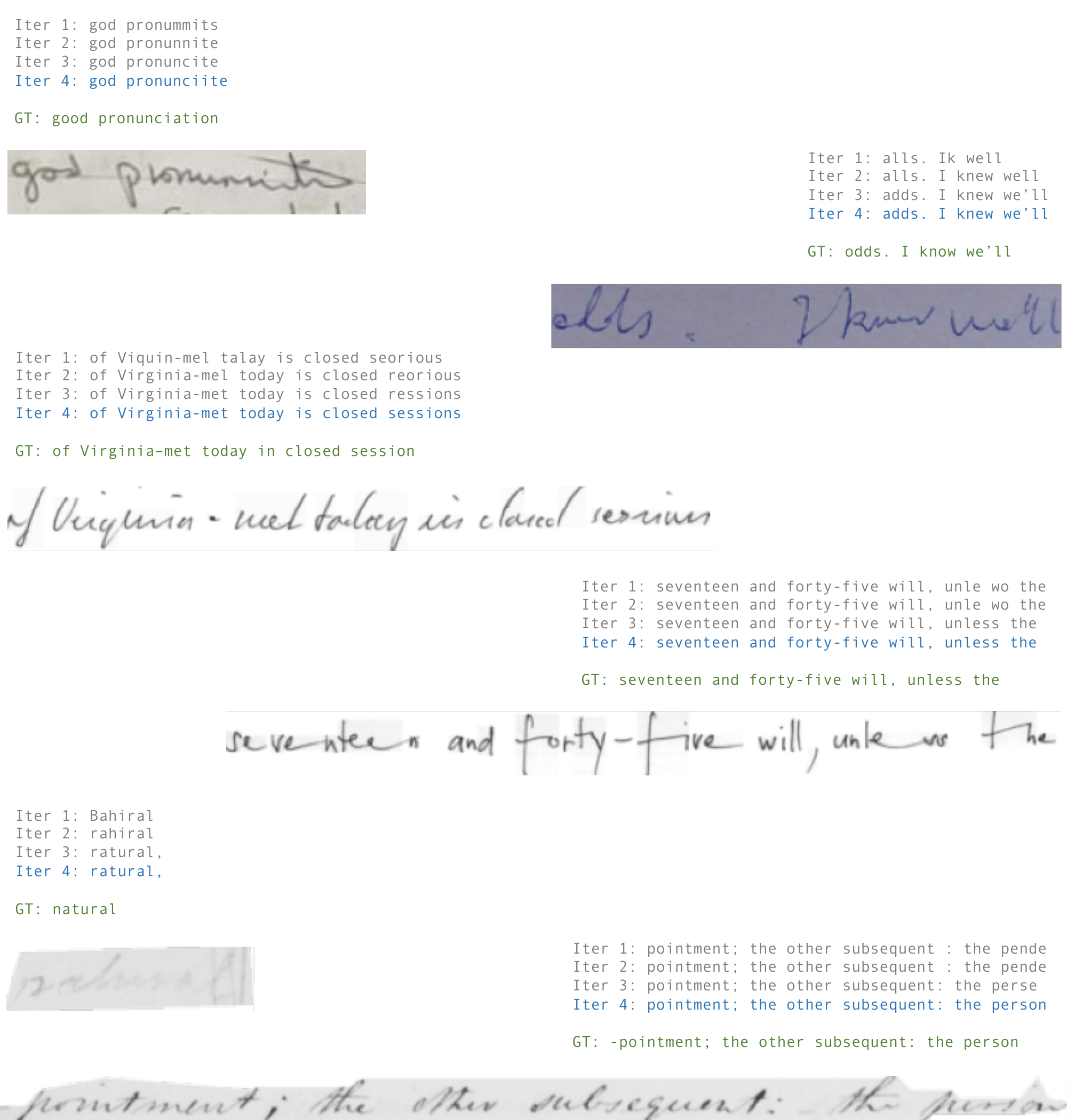}
    \caption{Exemplars showing the predictions over self-training iterations on two images each from the three datasets.}
    \label{fig:qualitative}
\end{figure*}

\section{Exemplars of predictions over progressive self-training iterations}
Figure \ref{fig:qualitative} shows the predictions after every iteration of the proposed self-training approach. As we can see, the predictions improve over iterations. For example, for the second one from top, the model quickly figures out that the style of `l' looks different from that of `d', and hence it corrects that in the later iterations. Similarly for the third example from the top, the multiple `s' in ``session" get correctly recognized over iterations. We see similar improvements in the other examples as well.

% \input{./tables/comp_baselines_kohtd}

%ICDAR
% Please add the following required packages to your document preamble:
% \usepackage[table,xcdraw]{xcolor}
% If you use beamer only pass "xcolor=table" option, i.e. \documentclass[xcolor=table]{beamer}
\begin{table*}
\centering
\resizebox{\textwidth}{!}{
\begin{tabular}{lll|ll|rrrrrrrrrrrrrrrrrrrr|c}
\toprule
 &
  &
  &
  \multicolumn{2}{|c|}{\textbf{Confidence}} &
  \multicolumn{20}{c|}{\textbf{Performance}} &
  \multicolumn{1}{l}{} \\
  \midrule
\multicolumn{1}{c}{\textbf{\begin{tabular}[c]{@{}c@{}}Update\\ self-labels\end{tabular}}} &
  \multicolumn{1}{c}{\textbf{Progressive}} &
  \multicolumn{1}{c}{\textbf{Weighting}} &
  \multicolumn{1}{|c}{\textbf{CTC}} &
  \multicolumn{1}{c|}{\textbf{NED}} &
  \multicolumn{1}{c}{\rotatebox[origin=c]{90}{\textbf{original}}} &
  \multicolumn{1}{c}{\rotatebox[origin=c]{90}{\textbf{brightness}}} &
  \multicolumn{1}{c}{\rotatebox[origin=c]{90}{\textbf{contrast}}} &
  \multicolumn{1}{c}{\rotatebox[origin=c]{90}{\textbf{defocus blur}}} &
  \multicolumn{1}{c}{\rotatebox[origin=c]{90}{\textbf{elastic}}} &
  \multicolumn{1}{c}{\rotatebox[origin=c]{90}{\textbf{fog}}} &
  \multicolumn{1}{c}{\rotatebox[origin=c]{90}{\textbf{frost}}} &
  \multicolumn{1}{c}{\rotatebox[origin=c]{90}{\textbf{gaussian blur}}} &
  \multicolumn{1}{c}{\rotatebox[origin=c]{90}{\textbf{gaussian noise}}} &
  \multicolumn{1}{c}{\rotatebox[origin=c]{90}{\textbf{glass blur}}} &
  \multicolumn{1}{c}{\rotatebox[origin=c]{90}{\textbf{impulse noise}}} &
  \multicolumn{1}{c}{\rotatebox[origin=c]{90}{\textbf{jpeg}}} &
  \multicolumn{1}{c}{\rotatebox[origin=c]{90}{\textbf{motion blur}}} &
  \multicolumn{1}{c}{\rotatebox[origin=c]{90}{\textbf{pixelate}}} &
  \multicolumn{1}{c}{\rotatebox[origin=c]{90}{\textbf{saturate}}} &
  \multicolumn{1}{c}{\rotatebox[origin=c]{90}{\textbf{shot noise}}} &
  \multicolumn{1}{c}{\rotatebox[origin=c]{90}{\textbf{snow}}} &
  \multicolumn{1}{c}{\rotatebox[origin=c]{90}{\textbf{spatter}}} &
  \multicolumn{1}{c}{\rotatebox[origin=c]{90}{\textbf{speckle noise}}} &
  \multicolumn{1}{c|}{\rotatebox[origin=c]{90}{\textbf{zoom blur}}} &
  \multicolumn{1}{c}{\textbf{AVERAGE}} \\
  \midrule
 &
  & 
  & 
  &
  &
  14.4 &
  14.3 &
  15.6 &
  39.2 &
  \textbf{42.7} &
  16.2 &
  29.6 &
  19.9 &
  26.3 &
  35.5 &
  36.9 &
  27.0 &
  31.0 &
  26.2 &
  14.4 &
  35.8 &
  23.3 &
  15.9 &
  37.9 &
  36.4 &
  26.9 \\
\multicolumn{1}{c} \checkmark &
  &
  &
  &
  &
  13.3 &
  13.2 &
  \textbf{14.1} &
  40.0 &
  42.8 &
  15.7 &
  28.8 &
  \textbf{18.1} &
  23.6 &
  31.3 &
  32.2 &
  23.2 &
  27.6 &
  20.9 &
  13.7 &
  32.3 &
  21.8 &
  14.7 &
  34.8 &
  \textbf{36.3} &
  24.9 \\
\multicolumn{1}{c} \checkmark &
  \multicolumn{1}{c} \checkmark &
  &
  \multicolumn{1}{c} \checkmark &
  &
  13.5 &
  13.5 &
  14.7 &
  39.8 &
  43.5 &
  15.2 &
  29.4 &
  18.8 &
  24.0 &
  32.1 &
  33.1 &
  23.7 &
  28.5 &
  21.5 &
  13.5 &
  32.8 &
  22.0 &
  15.0 &
  35.3 &
  36.8 &
  25.3 \\
\multicolumn{1}{c} \checkmark &
  \multicolumn{1}{c} \checkmark &
  \multicolumn{1}{c} \checkmark &
  \multicolumn{1}{c} \checkmark &
  &
  13.5 &
  13.4 &
  14.7 &
  39.1 &
  47.2 &
  15.2 &
  29.3 &
  18.6 &
  24.0 &
  32.1 &
  32.9 &
  23.7 &
  28.6 &
  21.3 &
  13.5 &
  33.0 &
  21.8 &
  15.0 &
  35.1 &
  37.4 &
  25.5 \\
\multicolumn{1}{c} \checkmark &
  \multicolumn{1}{c} \checkmark &
  &
  &
  \multicolumn{1}{c} \checkmark &
  13.3 &
  13.2 &
  14.5 &
  38.9 &
  43.4 &
  15.1 &
  28.8 &
  18.4 &
  23.4 &
  30.5 &
  31.9 &
  23.0 &
  28.0 &
  20.7 &
  13.3 &
  31.8 &
  21.7 &
  14.8 &
  33.8 &
  37.3 &
  24.8 \\
\multicolumn{1}{c} \checkmark &
  \multicolumn{1}{c} \checkmark &
  \multicolumn{1}{c} \checkmark &
  &
  \multicolumn{1}{c} \checkmark &
  \textbf{13.2} &
  \textbf{13.0} &
  14.2 &
  \textbf{39.1} &
  46.8 &
  \textbf{14.9} &
  \textbf{28.4} &
  18.2 &
  \textbf{22.8} &
  \textbf{30.5} &
  \textbf{31.3} &
  \textbf{22.6} &
  \textbf{27.9} &
  \textbf{20.1} &
  \textbf{13.1} &
  \textbf{31.4} &
  \textbf{21.0} &
  \textbf{14.6} &
  \textbf{33.4} &
  37.0 &
  \textbf{24.7} \\
  \bottomrule
\end{tabular}
}
\caption{Ablation results for different setting of self-training to update the optical model. The results are on ICDAR2015-HTR.}
\label{table:icdar_table}
\end{table*}

%GNHK
% Please add the following required packages to your document preamble:
% \usepackage[table,xcdraw]{xcolor}
% If you use beamer only pass "xcolor=table" option, i.e. \documentclass[xcolor=table]{beamer}
\begin{table*}
\centering
\resizebox{\textwidth}{!}{
\begin{tabular}{lll|ll|rrrrrrrrrrrrrrrrrrrr|c}
\toprule
 &
   &
   &
  \multicolumn{2}{|c|}{\textbf{Confidence}} &
  \multicolumn{20}{c|}{\textbf{Performance}} &
  \multicolumn{1}{l}{} \\
  \midrule
\multicolumn{1}{c}{\textbf{\begin{tabular}[c]{@{}c@{}}Update\\ self-labels\end{tabular}}} &
  \multicolumn{1}{c}{\textbf{Progressive}} &
  \multicolumn{1}{c}{\textbf{Weighting}} &
  \multicolumn{1}{|c}{\textbf{CTC}} &
  \multicolumn{1}{c|}{\textbf{NED}} &
  \multicolumn{1}{c}{\rotatebox[origin=c]{90}{\textbf{original}}} &
  \multicolumn{1}{c}{\rotatebox[origin=c]{90}{\textbf{brightness}}} &
  \multicolumn{1}{c}{\rotatebox[origin=c]{90}{\textbf{contrast}}} &
  \multicolumn{1}{c}{\rotatebox[origin=c]{90}{\textbf{defocus blur}}} &
  \multicolumn{1}{c}{\rotatebox[origin=c]{90}{\textbf{elastic}}} &
  \multicolumn{1}{c}{\rotatebox[origin=c]{90}{\textbf{fog}}} &
  \multicolumn{1}{c}{\rotatebox[origin=c]{90}{\textbf{frost}}} &
  \multicolumn{1}{c}{\rotatebox[origin=c]{90}{\textbf{gaussian blur}}} &
  \multicolumn{1}{c}{\rotatebox[origin=c]{90}{\textbf{gaussian noise}}} &
  \multicolumn{1}{c}{\rotatebox[origin=c]{90}{\textbf{glass blur}}} &
  \multicolumn{1}{c}{\rotatebox[origin=c]{90}{\textbf{impulse noise}}} &
  \multicolumn{1}{c}{\rotatebox[origin=c]{90}{\textbf{jpeg}}} &
  \multicolumn{1}{c}{\rotatebox[origin=c]{90}{\textbf{motion blur}}} &
  \multicolumn{1}{c}{\rotatebox[origin=c]{90}{\textbf{pixelate}}} &
  \multicolumn{1}{c}{\rotatebox[origin=c]{90}{\textbf{saturate}}} &
  \multicolumn{1}{c}{\rotatebox[origin=c]{90}{\textbf{shot noise}}} &
  \multicolumn{1}{c}{\rotatebox[origin=c]{90}{\textbf{snow}}} &
  \multicolumn{1}{c}{\rotatebox[origin=c]{90}{\textbf{spatter}}} &
  \multicolumn{1}{c}{\rotatebox[origin=c]{90}{\textbf{speckle noise}}} &
  \multicolumn{1}{c|}{\rotatebox[origin=c]{90}{\textbf{zoom blur}}} &
  \multicolumn{1}{c}{\textbf{AVERAGE}} \\
  \midrule
 &
   &
   &
   &
   &
  14.0 &
  14.5 &
  15.8 &
  \textbf{30.2} &
  \textbf{56.9} &
  17.3 &
  21.2 &
  22.1 &
  20.1 &
  38.4 &
  22.6 &
  18.4 &
  33.5 &
  20.2 &
  14.6 &
  22.7 &
  26.1 &
  18.9 &
  19.8 &
  29.7 &
  23.8 \\
\multicolumn{1}{c} \checkmark &
   &
   &
   &
   &
  13.9 &
  14.1 &
  15.4 &
  30.8 &
  57.5 &
  16.9 &
  20.8 &
  22.4 &
  19.3 &
  35.2 &
  21.8 &
  17.2 &
  30.3 &
  18.4 &
  14.4 &
  22.0 &
  25.0 &
  18.4 &
  19.1 &
  29.3 &
  23.1 \\
\multicolumn{1}{c} \checkmark &
  \multicolumn{1}{c} \checkmark &
   &
  \multicolumn{1}{c} \checkmark &
   &
  13.9 &
  14.3 &
  16.3 &
  31.6 &
  58.4 &
  17.1 &
  24.2 &
  28.8 &
  19.5 &
  35.2 &
  22.3 &
  17.5 &
  30.6 &
  18.5 &
  14.4 &
  22.3 &
  25.3 &
  18.5 &
  19.1 &
  28.9 &
  23.8 \\
\multicolumn{1}{c} \checkmark &
  \multicolumn{1}{c} \checkmark &
  \multicolumn{1}{c} \checkmark &
  \multicolumn{1}{c} \checkmark &
   &
  13.9 &
  14.3 &
  15.7 &
  31.6 &
  58.3 &
  17.2 &
  20.9 &
  23.0 &
  19.5 &
  34.8 &
  21.9 &
  17.5 &
  30.7 &
  18.5 &
  14.4 &
  22.0 &
  25.0 &
  18.5 &
  19.2 &
  28.6 &
  23.3 \\
\multicolumn{1}{c} \checkmark &
  \multicolumn{1}{c} \checkmark &
   &
   &
  \multicolumn{1}{c} \checkmark &
  13.8 &
  14.2 &
  15.6 &
  31.0 &
  58.1 &
  16.9 &
  20.9 &
  22.6 &
  19.2 &
  34.5 &
  21.6 &
  17.2 &
  30.0 &
  18.0 &
  14.2 &
  22.9 &
  25.0 &
  18.4 &
  19.0 &
  30.6 &
  23.2 \\
\multicolumn{1}{c} \checkmark &
  \multicolumn{1}{c} \checkmark &
  \multicolumn{1}{c} \checkmark &
   &
  \multicolumn{1}{c} \checkmark &
  \textbf{13.3} &
  \textbf{13.8} &
  \textbf{15.1} &
  30.7 &
  58.1 &
  \textbf{16.4} &
  \textbf{20.4} &
  \textbf{22.2} &
  \textbf{18.8} &
  \textbf{33.7} &
  \textbf{21.1} &
  \textbf{16.7} &
  \textbf{29.7} &
  \textbf{17.5} &
  \textbf{13.7} &
  \textbf{21.3} &
  \textbf{24.2} &
  \textbf{17.8} &
  \textbf{18.5} &
  \textbf{28.0} &
  \textbf{22.6} \\
  \bottomrule
  \end{tabular}
}
\caption{Ablation results for different setting of self-training to update the optical model. The results are on GNHK.}
\label{table:gnhk_table}
\end{table*}
%IAM
% Please add the following required packages to your document preamble:
% \usepackage[table,xcdraw]{xcolor}
% If you use beamer only pass "xcolor=table" option, i.e. \documentclass[xcolor=table]{beamer}
\begin{table*}
\centering
\resizebox{\textwidth}{!}{
\begin{tabular}{lll|ll|rrrrrrrrrrrrrrrrrrrr|c}
\toprule
 &
   &
   &
  \multicolumn{2}{|c|}{\textbf{Confidence}} &
  \multicolumn{20}{c|}{\textbf{Performance}} &
  \multicolumn{1}{l}{} \\
  \midrule
\multicolumn{1}{c}{\textbf{\begin{tabular}[c]{@{}c@{}}Update\\ self-labels\end{tabular}}} &
  \multicolumn{1}{c}{\textbf{Progressive}} &
  \multicolumn{1}{c}{\textbf{Weighting}} &
  \multicolumn{1}{|c}{\textbf{CTC}} &
  \multicolumn{1}{c|}{\textbf{NED}} &
  \multicolumn{1}{c}{\rotatebox[origin=c]{90}{\textbf{original}}} &
  \multicolumn{1}{c}{\rotatebox[origin=c]{90}{\textbf{brightness}}} &
  \multicolumn{1}{c}{\rotatebox[origin=c]{90}{\textbf{contrast}}} &
  \multicolumn{1}{c}{\rotatebox[origin=c]{90}{\textbf{defocus blur}}} &
  \multicolumn{1}{c}{\rotatebox[origin=c]{90}{\textbf{elastic}}} &
  \multicolumn{1}{c}{\rotatebox[origin=c]{90}{\textbf{fog}}} &
  \multicolumn{1}{c}{\rotatebox[origin=c]{90}{\textbf{frost}}} &
  \multicolumn{1}{c}{\rotatebox[origin=c]{90}{\textbf{gaussian blur}}} &
  \multicolumn{1}{c}{\rotatebox[origin=c]{90}{\textbf{gaussian noise}}} &
  \multicolumn{1}{c}{\rotatebox[origin=c]{90}{\textbf{glass blur}}} &
  \multicolumn{1}{c}{\rotatebox[origin=c]{90}{\textbf{impulse noise}}} &
  \multicolumn{1}{c}{\rotatebox[origin=c]{90}{\textbf{jpeg}}} &
  \multicolumn{1}{c}{\rotatebox[origin=c]{90}{\textbf{motion blur}}} &
  \multicolumn{1}{c}{\rotatebox[origin=c]{90}{\textbf{pixelate}}} &
  \multicolumn{1}{c}{\rotatebox[origin=c]{90}{\textbf{saturate}}} &
  \multicolumn{1}{c}{\rotatebox[origin=c]{90}{\textbf{shot noise}}} &
  \multicolumn{1}{c}{\rotatebox[origin=c]{90}{\textbf{snow}}} &
  \multicolumn{1}{c}{\rotatebox[origin=c]{90}{\textbf{spatter}}} &
  \multicolumn{1}{c}{\rotatebox[origin=c]{90}{\textbf{speckle noise}}} &
  \multicolumn{1}{c|}{\rotatebox[origin=c]{90}{\textbf{zoom blur}}} &
  \multicolumn{1}{c}{\textbf{AVERAGE}} \\
  \midrule
 &
   &
   &
   &
   &
  6.0 &
  7.1 &
  7.1 &
  20.1 &
  57.7 &
  6.8 &
  11.5 &
  13.2 &
  7.4 &
  26.3 &
  11.8 &
  7.6 &
  22.4 &
  12.2 &
  6.1 &
  9.4 &
  13.7 &
  6.7 &
  9.1 &
  38.1 &
  15.0 \\
\multicolumn{1}{c} \checkmark &
   &
   &
   &
   &
  5.7 &
  6.6 &
  6.4 &
  19.9 &
  58.6 &
  6.2 &
  11.1 &
  12.4 &
  6.8 &
  19.5 &
  10.6 &
  6.9 &
  16.7 &
  8.6 &
  5.8 &
  8.6 &
  12.8 &
  6.3 &
  8.4 &
  38.3 &
  13.8 \\
\multicolumn{1}{c} \checkmark &
  \multicolumn{1}{c} \checkmark &
   &
  \multicolumn{1}{c} \checkmark &
   &
  5.9 &
  6.8 &
  6.9 &
  20.0 &
  \textbf{57.7} &
  6.3 &
  11.0 &
  13.2 &
  7.3 &
  26.2 &
  11.2 &
  7.0 &
  18.2 &
  8.5 &
  5.8 &
  8.8 &
  12.9 &
  6.5 &
  8.4 &
  38.0 &
  14.3 \\
\multicolumn{1}{c} \checkmark &
  \multicolumn{1}{c} \checkmark &
  \multicolumn{1}{c} \checkmark &
  \multicolumn{1}{c} \checkmark &
   &
  5.8 &
  6.8 &
  6.7 &
  20.7 &
  59.5 &
  6.5 &
  11.5 &
  13.2 &
  7.0 &
  20.1 &
  10.7 &
  7.1 &
  17.7 &
  8.9 &
  5.9 &
  8.9 &
  13.2 &
  6.5 &
  8.6 &
  38.5 &
  14.2 \\
\multicolumn{1}{c} \checkmark &
  \multicolumn{1}{c} \checkmark &
   &
   &
  \multicolumn{1}{c} \checkmark &
  5.7 &
  6.6 &
  6.5 &
  19.9 &
  59.1 &
  6.1 &
  11.2 &
  12.7 &
  6.9 &
  19.9 &
  10.2 &
  6.8 &
  16.9 &
  8.5 &
  5.7 &
  8.8 &
  12.9 &
  6.3 &
  8.3 &
  37.5 &
  13.8 \\
\multicolumn{1}{c} \checkmark &
  \multicolumn{1}{c} \checkmark &
  \multicolumn{1}{c} \checkmark &
   &
  \multicolumn{1}{c} \checkmark &
  \textbf{5.4} &
  \textbf{6.4} &
  \textbf{6.2} &
  \textbf{19.7} &
  58.7 &
  \textbf{5.9} &
  \textbf{10.9} &
  \textbf{12.4} &
  \textbf{6.4} &
  \textbf{19.2} &
  \textbf{9.9} &
  \textbf{6.4} &
  \textbf{16.6} &
  \textbf{8.0} &
  \textbf{5.4} &
  \textbf{8.0} &
  \textbf{12.5} &
  \textbf{5.9} &
  \textbf{7.9} &
  \textbf{35.6} &
  \textbf{13.4} \\
  \bottomrule
\end{tabular}
}
\caption{Ablation results for different setting of self-training to update the optical model. The results are on IAM.}
\label{table:iam_table}
\end{table*}
% CVL
% Please add the following required packages to your document preamble:
% \usepackage[table,xcdraw]{xcolor}
% If you use beamer only pass "xcolor=table" option, i.e. \documentclass[xcolor=table]{beamer}
\begin{table*}
\centering
\resizebox{\textwidth}{!}{
\begin{tabular}{lll|ll|rrrrrrrrrrrrrrrrrrrr|c}
\toprule
 &
  &
  &
  \multicolumn{2}{|c|}{\textbf{Confidence}} &
  \multicolumn{20}{c|}{\textbf{Performance}} &
  \multicolumn{1}{l}{} \\
  \midrule
\multicolumn{1}{c}{\textbf{\begin{tabular}[c]{@{}c@{}}Update\\ self-labels\end{tabular}}} &
  \multicolumn{1}{c}{\textbf{Progressive}} &
  \multicolumn{1}{c}{\textbf{Weighting}} &
  \multicolumn{1}{|c}{\textbf{CTC}} &
  \multicolumn{1}{c|}{\textbf{NED}} &
  \multicolumn{1}{c}{\rotatebox[origin=c]{90}{\textbf{original}}} &
  \multicolumn{1}{c}{\rotatebox[origin=c]{90}{\textbf{brightness}}} &
  \multicolumn{1}{c}{\rotatebox[origin=c]{90}{\textbf{contrast}}} &
  \multicolumn{1}{c}{\rotatebox[origin=c]{90}{\textbf{defocus blur}}} &
  \multicolumn{1}{c}{\rotatebox[origin=c]{90}{\textbf{elastic}}} &
  \multicolumn{1}{c}{\rotatebox[origin=c]{90}{\textbf{fog}}} &
  \multicolumn{1}{c}{\rotatebox[origin=c]{90}{\textbf{frost}}} &
  \multicolumn{1}{c}{\rotatebox[origin=c]{90}{\textbf{gaussian blur}}} &
  \multicolumn{1}{c}{\rotatebox[origin=c]{90}{\textbf{gaussian noise}}} &
  \multicolumn{1}{c}{\rotatebox[origin=c]{90}{\textbf{glass blur}}} &
  \multicolumn{1}{c}{\rotatebox[origin=c]{90}{\textbf{impulse noise}}} &
  \multicolumn{1}{c}{\rotatebox[origin=c]{90}{\textbf{jpeg}}} &
  \multicolumn{1}{c}{\rotatebox[origin=c]{90}{\textbf{motion blur}}} &
  \multicolumn{1}{c}{\rotatebox[origin=c]{90}{\textbf{pixelate}}} &
  \multicolumn{1}{c}{\rotatebox[origin=c]{90}{\textbf{saturate}}} &
  \multicolumn{1}{c}{\rotatebox[origin=c]{90}{\textbf{shot noise}}} &
  \multicolumn{1}{c}{\rotatebox[origin=c]{90}{\textbf{snow}}} &
  \multicolumn{1}{c}{\rotatebox[origin=c]{90}{\textbf{spatter}}} &
  \multicolumn{1}{c}{\rotatebox[origin=c]{90}{\textbf{speckle noise}}} &
  \multicolumn{1}{c|}{\rotatebox[origin=c]{90}{\textbf{zoom blur}}} &
  \multicolumn{1}{c}{\textbf{AVERAGE}} \\
  \midrule
 &
  & 
  & 
  &
  &
  8.6 &
  11.0 &
  12.6 &
  30.4 &
  \textbf{59.0} &
  11.5 &
  \textbf{37.0} &
  \textbf{20.6} &
  15.4 &
  34.1 &
  27.9 &
  14.7 &
  35.8 &
  16.6 &
  12.9 &
  26.4 &
  51.8 &
  11.5 &
  26.6 &
  \textbf{46.6} &
  25.5 \\
\multicolumn{1}{c} \checkmark &
  &
  &
  &
  &
  8.1 &
  10.6 &
  12.0 &
  37.3 &
  62.4 &
  11.0 &
  48.7 &
  24.2 &
  18.8 &
  28.8 &
  25.6 &
  13.1 &
  30.6 &
  13.3 &
  12.7 &
  \textbf{23.1} &
  \textbf{51.8} &
  11.1 &
  \textbf{23.6} &
  52.9 &
  26.0 \\
\multicolumn{1}{c} \checkmark &
  \multicolumn{1}{c} \checkmark &
  &
  \multicolumn{1}{c} \checkmark &
  &
  8.5 &
  10.9 &
  12.4 &
  34.7 &
  61.8 &
  11.4 &
  39.7 &
  24.0 &
  14.8 &
  29.6 &
  25.8 &
  13.3 &
  31.1 &
  13.6 &
  13.6 &
  24.8 &
  53.4 &
  11.3 &
  25.3 &
  55.1 &
  25.8  \\
\multicolumn{1}{c} \checkmark &
  \multicolumn{1}{c} \checkmark &
  \multicolumn{1}{c} \checkmark &
  \multicolumn{1}{c} \checkmark &
  &
  8.5 &
  11.3 &
  12.5 &
  35.7 &
  62.6 &
  11.4 &
  39.7 &
  23.7 &
  14.9 &
  29.6 &
  26.4 &
  13.3 &
  31.0 &
  13.8 &
  13.3 &
  25.0 &
  53.2 &
  11.4 &
  25.4 &
  57.4 &
  26.0 \\
\multicolumn{1}{c} \checkmark &
  \multicolumn{1}{c} \checkmark &
  &
  &
  \multicolumn{1}{c} \checkmark &
  8.3 &
  10.7 &
  12.2 &
  34.4 &
  61.4 &
  11.1 &
  39.4 &
  23.3 &
  14.5 &
  29.4 &
  25.4 &
  13.0 &
  30.4 &
  13.1 &
  13.3 &
  24.4 &
  53.6 &
  11.2 &
  24.9 &
  55.4 &
  25.5 \\
\multicolumn{1}{c} \checkmark &
  \multicolumn{1}{c} \checkmark &
  \multicolumn{1}{c} \checkmark &
   &
  \multicolumn{1}{c} \checkmark &
  \textbf{7.8} &
  \textbf{10.2} &
  \textbf{11.5} &
  \textbf{33.8} &
  61.5 &
  \textbf{10.5} &
  39.0 &
  22.8 &
  \textbf{13.9} &
  \textbf{29.1} &
  \textbf{24.8} &
  \textbf{12.3} &
  \textbf{30.3} &
  \textbf{12.4} &
  \textbf{12.6} &
  23.9 &
  53.4 &
  \textbf{10.5} &
  24.4 &
  56.2 &
  \textbf{25.0} \\
  \bottomrule
  \end{tabular}
}
\caption{Ablation for different setting of self-training to update the optical model. The results are on CVL.}
\label{table:cvl_table}
\end{table*}

% KOHTD
% Please add the following required packages to your document preamble:
% \usepackage[table,xcdraw]{xcolor}
% If you use beamer only pass "xcolor=table" option, i.e. \documentclass[xcolor=table]{beamer}
\begin{table*}
\centering
\resizebox{\textwidth}{!}{
\begin{tabular}{lll|ll|rrrrrrrrrrrrrrrrrrrr|c}
\toprule
 &
  &
  &
  \multicolumn{2}{|c|}{\textbf{Confidence}} &
  \multicolumn{20}{c|}{\textbf{Performance}} &
  \multicolumn{1}{l}{} \\
  \midrule
\multicolumn{1}{c}{\textbf{\begin{tabular}[c]{@{}c@{}}Update\\ self-labels\end{tabular}}} &
  \multicolumn{1}{c}{\textbf{Progressive}} &
  \multicolumn{1}{c}{\textbf{Weighting}} &
  \multicolumn{1}{|c}{\textbf{CTC}} &
  \multicolumn{1}{c|}{\textbf{NED}} &
  \multicolumn{1}{c}{\rotatebox[origin=c]{90}{\textbf{original}}} &
  \multicolumn{1}{c}{\rotatebox[origin=c]{90}{\textbf{brightness}}} &
  \multicolumn{1}{c}{\rotatebox[origin=c]{90}{\textbf{contrast}}} &
  \multicolumn{1}{c}{\rotatebox[origin=c]{90}{\textbf{defocus blur}}} &
  \multicolumn{1}{c}{\rotatebox[origin=c]{90}{\textbf{elastic}}} &
  \multicolumn{1}{c}{\rotatebox[origin=c]{90}{\textbf{fog}}} &
  \multicolumn{1}{c}{\rotatebox[origin=c]{90}{\textbf{frost}}} &
  \multicolumn{1}{c}{\rotatebox[origin=c]{90}{\textbf{gaussian blur}}} &
  \multicolumn{1}{c}{\rotatebox[origin=c]{90}{\textbf{gaussian noise}}} &
  \multicolumn{1}{c}{\rotatebox[origin=c]{90}{\textbf{glass blur}}} &
  \multicolumn{1}{c}{\rotatebox[origin=c]{90}{\textbf{impulse noise}}} &
  \multicolumn{1}{c}{\rotatebox[origin=c]{90}{\textbf{jpeg}}} &
  \multicolumn{1}{c}{\rotatebox[origin=c]{90}{\textbf{motion blur}}} &
  \multicolumn{1}{c}{\rotatebox[origin=c]{90}{\textbf{pixelate}}} &
  \multicolumn{1}{c}{\rotatebox[origin=c]{90}{\textbf{saturate}}} &
  \multicolumn{1}{c}{\rotatebox[origin=c]{90}{\textbf{shot noise}}} &
  \multicolumn{1}{c}{\rotatebox[origin=c]{90}{\textbf{snow}}} &
  \multicolumn{1}{c}{\rotatebox[origin=c]{90}{\textbf{spatter}}} &
  \multicolumn{1}{c}{\rotatebox[origin=c]{90}{\textbf{speckle noise}}} &
  \multicolumn{1}{c|}{\rotatebox[origin=c]{90}{\textbf{zoom blur}}} &
  \multicolumn{1}{c}{\textbf{AVERAGE}} \\
  \midrule
 &
  & 
  & 
  &
  &
  23.3 &
  27.2 &
  27.9 &
  36.5 &
  29.0 &
  31.8 &
  55.2 &
  26.9 &
  26.9 &
  31.5 &
  33.8 &
  29.1 &
  39.0 &
  27.0 &
  24.9 &
  30.1 &
  43.5 &
  25.0 &
  31.8 &
  30.5 &
  31.5 \\
\multicolumn{1}{c} \checkmark &
  &
  &
  &
  &
  20.1 &
  24.2 &
  24.7 &
  36.2 &
  28.6 &
  31.8 &
  56.4 &
  23.9 &
  23.8 &
  25.4 &
  29.2 &
  29.1 &
  34.0 &
  27.0 &
  21.9 &
  27.4 &
  41.2 &
  22.6 &
  25.4 &
  30.1 &
  28.8 \\
\multicolumn{1}{c} \checkmark &
  \multicolumn{1}{c} \checkmark &
  &
  \multicolumn{1}{c} \checkmark &
  &
  20.3 &
  24.7 &
  25.2 &
  32.6 &
  27.3 &
  31.9 &
  56.7 &
  23.9 &
  26.1 &
  25.8 &
  29.2 &
  28.6 &
  33.1 &
  26.6 &
  19.7 &
  28.3 &
  39.3 &
  22.3 &
  27.2 &
  32.1 &
  29.0 \\
\multicolumn{1}{c} \checkmark &
  \multicolumn{1}{c} \checkmark &
  \multicolumn{1}{c} \checkmark &
  \multicolumn{1}{c} \checkmark &
  &
  20.7 &
  24.9 &
  25.5 &
  25.4 &
  27.5 &
  31.9 &
  56.9 &
  23.5 &
  26.1 &
  25.6 &
  28.9 &
  28.6 &
  33.2 &
  26.3 &
  19.2 &
  28.1 &
  39.2 &
  22.5 &
  27.2 &
  32.8 &
  29.0 \\
\multicolumn{1}{c} \checkmark &
  \multicolumn{1}{c} \checkmark &
  &
  &
  \multicolumn{1}{c} \checkmark &
  18.7 &
  21.5 &
  22.7 &
  34.4 &
  27.5 &
  26.9 &
  55.7 &
  20.3 &
  23.1 &
  25.6 &
  27.8 &
  23.5 &
  33.2 &
  20.5 &
  18.8 &
  18.3 &
  39.0 &
  20.4 &
  26.1 &
  25.6 &
  26.8 \\
\multicolumn{1}{c} \checkmark &
  \multicolumn{1}{c} \checkmark &
  \multicolumn{1}{c} \checkmark &
   &
  \multicolumn{1}{c} \checkmark &
  \textbf{16.8} &
  \textbf{19.9} &
  \textbf{20.1} &
  \textbf{33.4} &
  \textbf{22.3} &
  \textbf{24.0} &
  \textbf{54.8} &
  \textbf{19.9} &
  \textbf{19.7} &
  \textbf{22.7} &
  \textbf{27.6} &
  \textbf{20.3} &
  \textbf{31.3} &
  \textbf{18.7} &
  \textbf{17.4} &
  \textbf{23.1} &
  \textbf{38.5} &
  \textbf{17.7} &
  \textbf{25.3} &
  \textbf{22.1} &
  \textbf{24.8} \\
  \bottomrule
  \end{tabular}
}
\caption{Ablation for different setting of self-training to update the optical model. The results are on KOHTD.}
\label{table:kohtd_table}
\end{table*}

\section{Category-wise ablation of self-training the optical model}
In Tables \ref{table:icdar_table}, \ref{table:gnhk_table}, \ref{table:iam_table}, \ref{table:cvl_table} and \ref{table:kohtd_table}, we show ablation results for various choices of our self-training approach. For ICDAR2015-HTR, we observe that our method performs better than the other options for most corruptions. Our augmentation based reweighting strategy with progressive update of labels shows CER improvements of as much as $\textbf{5.6\%}$ (for impulse noise corruption). %The elastic transformation contracts or stretches portions of a line image \cite{hendrycks2019benchmarking}, which leads to large degradation in image quality at the character level. Thus, errors in pseudo-labels generated, can propagate during the self-train process to further degrade the quality of predictions.

For the GNHK dataset, we observe similar trends as ICDAR2015-HTR. We observe an improvement in CER of as much as $\textbf{4.7\%}$ for the glass blur corruption. On an average, our proposed augmentation based TTA with reweighting performs the best and shows an improvement of $1.3\%$ over the baseline for all corruptions, including the original dataset.

In case of IAM, we observe that our method performs much better than the other options on average. Our method perform the best for majority of the corruptions, and shows an improvement of as much as $\textbf{7.1}\%$ (for the glass blur corruption). For a few of the corruptions adapting the model using predictions over all lines of an image as pseudo labels helps better. We speculate this happens because IAM being a much easier dataset, the predictions generated from the source model itself are of good quality, and hence serve as good self-labels. Thus, updating the model with the self-labels for all lines, over all iterations performs a bit better.

For the CVL dataset, we observe improvements in most cases, with a maximum improvement of $\textbf{5.5}$\%, except a few corruptions like defocus\_blur, elastic, gaussian\_blur and zoom\_blur. This can be attributed to the correctness of the self-labels due to the hard inherent distortions these corruptions introduce. 

In case of the dataset, KOHTD, we observe a consistent improvement across all corruptions. We observe that our progressive update of self-labels with reweighting shows CER improvements of at least $0.3\%$ and as much as $\textbf{8.8}\%$. KOHTD, being a low resource language dataset, we believe that the divergence controlled adaptation we introduce helps identify and learn from the writer's idiosyncrasies very effectively.

% \section{Results mentioned for ICDAR only in the main paper}
% Table 2, 3, 4 and Figures 6 and 7 for each GNHK and IAM as well. 

\section{Understanding the correlation between our confidence metric and correctness.}
\begin{figure*}
    \centering
    \includegraphics[scale=0.25]{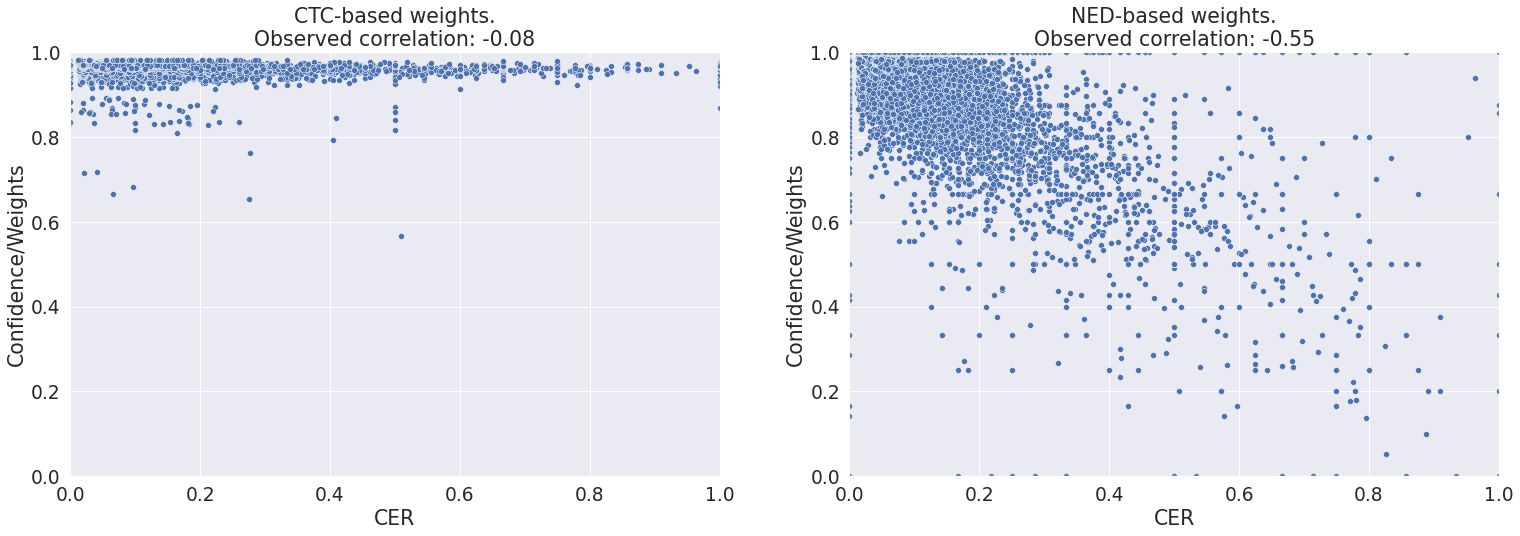}
    \caption{\small \textbf{Correlation between correctness and confidence metric used on ICDAR2015-HTR} Plot of ground-truth CER vs the confidence obtained based on the CTC loss (left), and the confidence obtained using our augmentation based confidence metric (right). Our confidence metric assigns higher values to the lines with lower CER. Whereas, for the same CTC loss based confidence, we observe a wide range of CER, indicating a low correlation with correctness.}
    \label{fig:icdar_main_plot}
\end{figure*}

\begin{figure*}
    \centering
    \includegraphics[scale=0.25]{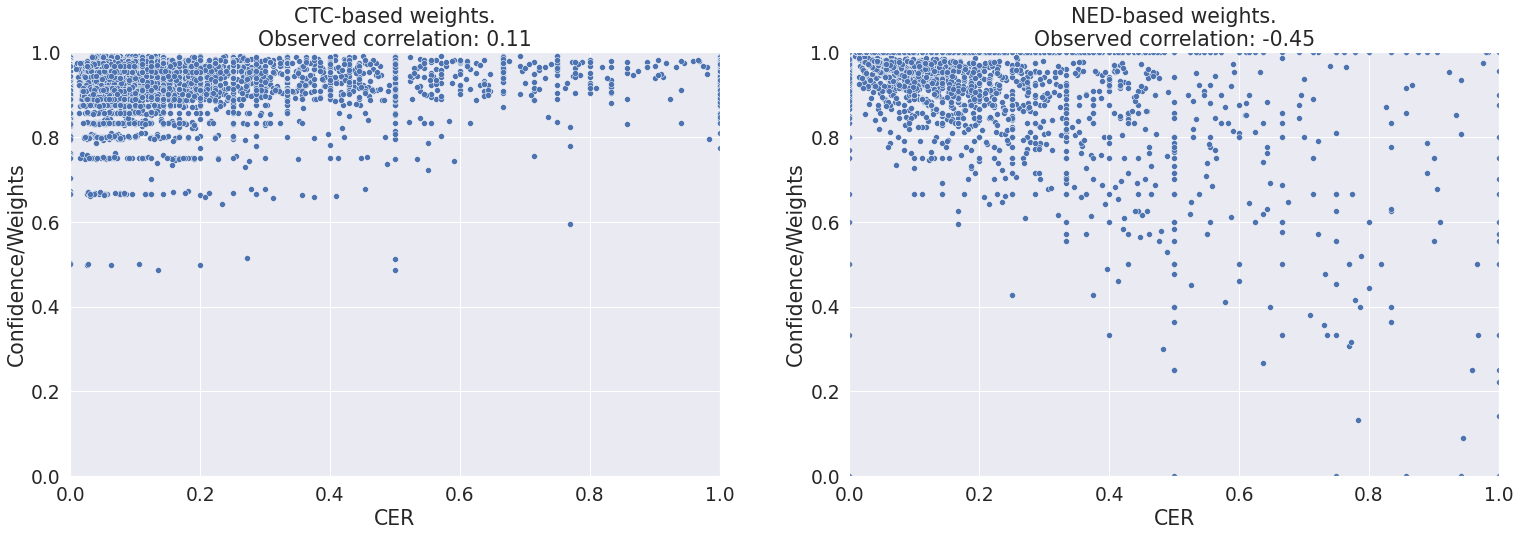}
    \caption{\small \textbf{Correlation between correctness and confidence metric used on GNHK} Plot of ground-truth CER vs the confidence obtained based on the CTC loss (left), and the confidence obtained using our augmentation based confidence metric (right). Our confidence metric assigns higher values to the lines with lower CER. Whereas, for the same CTC loss based confidence, high weights are assigned to the lines irrespective of their CERs, indicating a low correlation with correctness. There is even a positve correlation between the confidence using CTC loss and the CER, which should have been otherwise negative, as more confidence should be assigned to lines with lower CER, i.e., ones which are supposedly less wrong.}
    \label{fig:gnhk_main_plot}
\end{figure*}

\begin{figure*}
    \centering
    \includegraphics[scale=0.25]{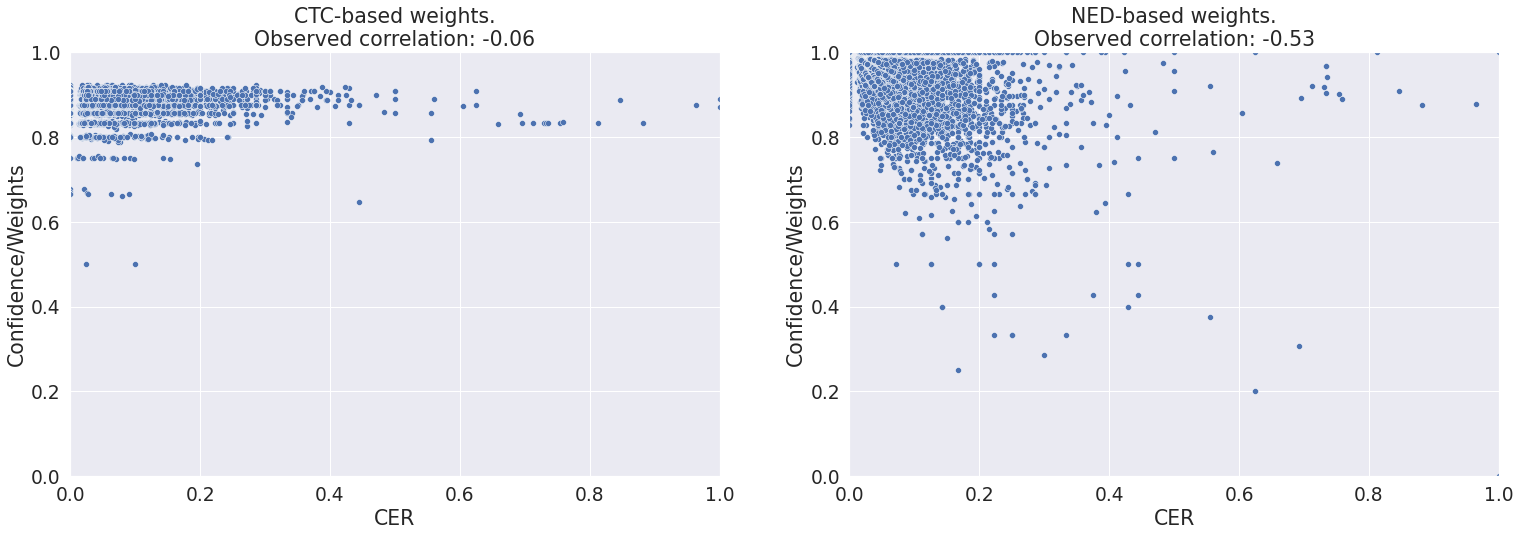}
    \caption{\small \textbf{Correlation between correctness and confidence metric used on IAM} Plot of ground-truth CER vs the confidence obtained based on the CTC loss (left), and the confidence obtained using our augmentation based confidence metric (right). Our confidence metric assigns higher values to the lines with lower CER. Whereas, for the same CTC loss based confidence, high weights are assigned to the lines irrespective of their CERs, indicating a low correlation with correctness.}
    \label{fig:iam_main_plot}
\end{figure*}

\begin{figure*}
    \centering
    \includegraphics[scale=0.25]{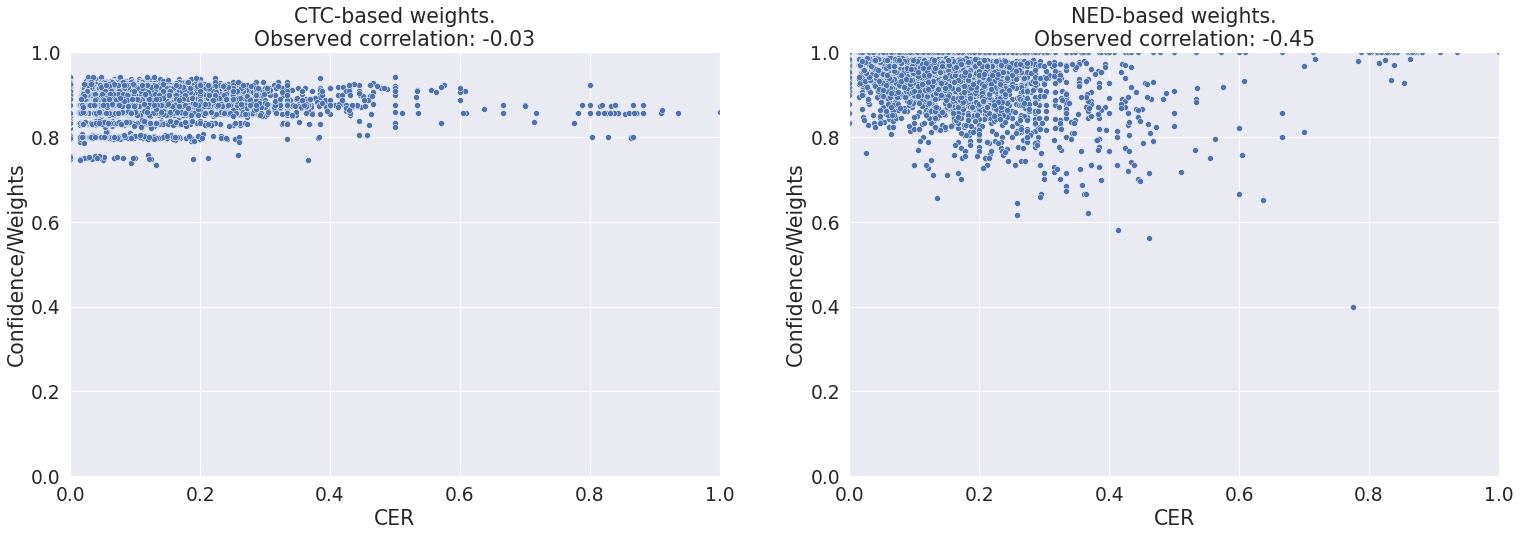}
    \caption{\small \textbf{Correlation between correctness and confidence metric used on CVL} Plot of ground-truth CER vs the confidence obtained based on the CTC loss (left), and the confidence obtained using our augmentation based confidence metric (right). The CTC based metric assigns high confidence values to even those lines with high CER. Our CER based metric is more closely correlated with the error, as observed.}
    \label{fig:cvl_main_plot}
\end{figure*}

\begin{figure*}
    \centering
    \includegraphics[scale=0.25]{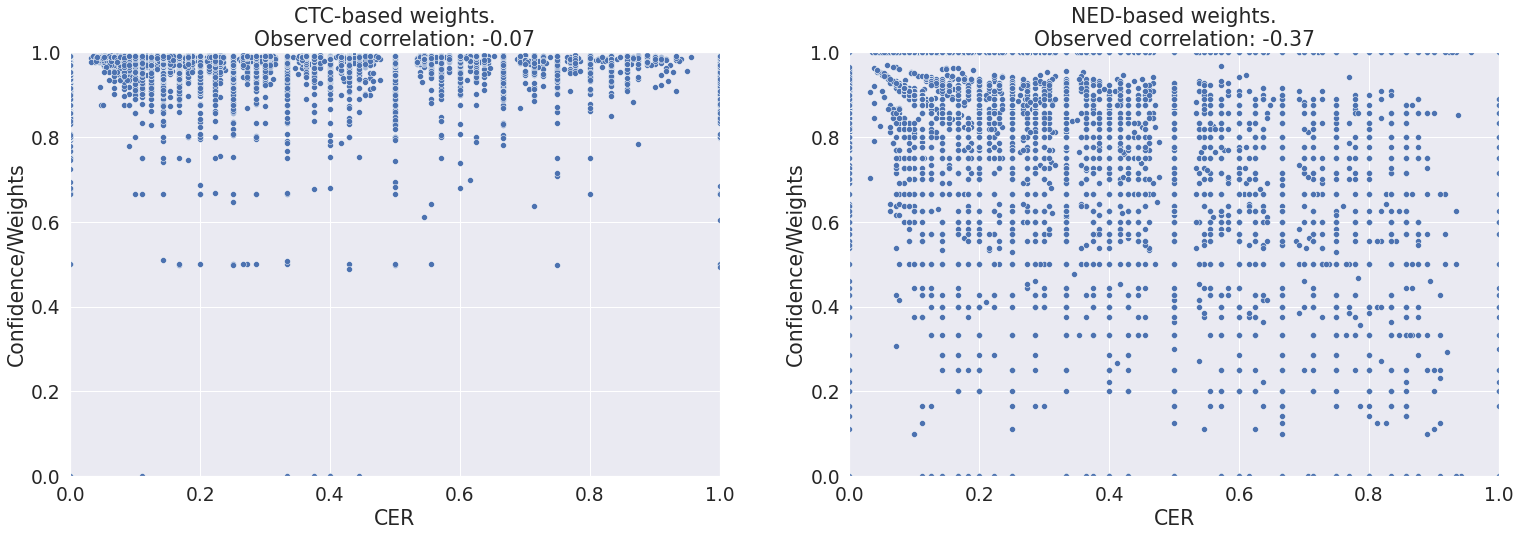}
    \caption{\small \textbf{Correlation between correctness and confidence metric used on KOHTD} Plot of ground-truth CER vs the confidence obtained based on the CTC loss (left), and the confidence obtained using our augmentation based confidence metric (right). The CTC based metric assigns high confidence to lines irrespective of their CERs. The unnatural blockyness of the plot is because the evaluation on this dataset is at a word level, and hence there is an inherent quantization of possible CER. }
    \label{fig:kohtd_main_plot}
\end{figure*}

In our algorithm, we use a confidence metric to both rank and weight losses for lines --  which forms the core of our progressive self-training approach. Our confidence metric (which we call the Normalized Edit Distance (NED) in the tables and figures) determines whether the optical model is smooth in the neighborhood of the input image. A baseline confidence metric is the CTC loss itself w.r.t. the top-1 prediction, with which we compare in Tables \ref{table:icdar_table}, \ref{table:gnhk_table}, \ref{table:iam_table}, \ref{table:cvl_table} and \ref{table:kohtd_table}. In order to better understand the usefulness of the proposed confidence metric, we find the correlation of the confidence metric and the CER of the self-labels with the ground-truth. We perform the analysis for both our metric and the CTC loss based weights (normalized over all lines in an image). The plots over the five datasets are shown in Figures \ref{fig:icdar_main_plot}, \ref{fig:gnhk_main_plot}, \ref{fig:iam_main_plot}, \ref{fig:cvl_main_plot} and \ref{fig:kohtd_main_plot}. As we can see from the plots on the right of the figures which represent the confidence scores derived using our augmentation based NED metric, as the CER for a line increases, the confidence of the model over the line also decreases. We see a much higher correlation between the CER and the proposed metric than the confidence based on the CTC loss.

%todo
\section{Ablation studies with varying number of iterations}
We evaluate our method for different number of iterations $K$ of our self-training algorithm. Recall, we progressively add $\frac{m}{K}$ lines to our set of confident lines, where $m$ is the total number of lines in an image. We run our algorithm for $ K \in \{1, 2, 4, 6, 8, 10, 12\}$. The results are shown for all corruptions in tables \ref{table:topx_icdar}, \ref{table:topx_gnhk}, \ref{table:topx_iam}, \ref{table:topx_cvl} and \ref{table:topx_kohtd}. As discussed in the paper, when we have very few update iterations, we do not allow the self-training algorithm to improve, and on the other hand, too many iterations may lead to divergence, as the algorithm does not use any ground-truth labels, but learns from self-labels only. We find that for all the datasets, number of iterations $K=4$, i.e., incrementally taking $25\%$ of the lines in each iteration results in the best performance. We can observe that choosing $K=4$ helps us strike the right balance between performance and total number of iterations.

%todo
\section{The Impact of LM on TTA}
\begin{figure*}
    \centering
    \includegraphics[scale=0.5]{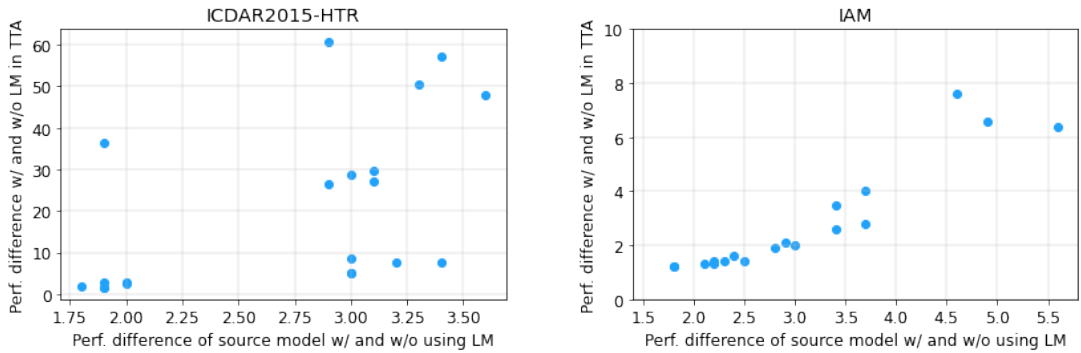}
    \caption{\small \textbf{The role of LM in adapting the optical model for ICDAR2015-HTR and IAM respectively.} Each point here denotes the relative improvement the LM introduces for various corrupted versions of the dataset. The x value of each point is the CER improvement when decoding is done using the LM Decoder vs GreedyDecoder when evaluating the source model. The y-axis shows the CER improvement when our proposed method is used with the LM Decoder vs GreedyDecoder.}
    \label{fig:lm_impact}
\end{figure*}

In the paper we show this analysis on GNHK, where we find that the performance significantly degrades if we do not use the LM in the self-training loop. This is because the LM acts as a cushion and prevents the model from diverging, which can otherwise happen because of the incorrect self-labels from just the optical model. Here, we show the analysis on two other datasets. When we do not use LM in the self-training algorithm, we observe a drop in performance from $23.9$ to $44.4$ on ICDAR2015-HTR, from $24.7$ to $31.4$ on GNHK, and from $13.6$ to $21$ on IAM. We also show a similar plot as in the paper in Figure \ref{fig:lm_impact} for ICDAR2015-HTR and IAM. The plot shows how the performance improvement offered by the LM on the source model inference can have an impact on the TTA performance. 


% todo: kohtd remaining
\section{Writer-specific Adaptation}
\begin{figure*}
     \centering
     \begin{subfigure}[b]{0.45\textwidth}
         \centering
         \includegraphics[width=\textwidth]{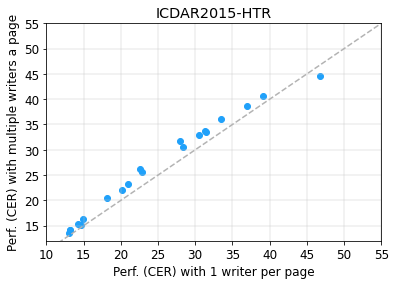}
         \caption{ICDAR2015-HTR}
         \label{fig:icdar_mw}
     \end{subfigure}
     \hfill
     \begin{subfigure}[b]{0.45\textwidth}
         \centering
         \includegraphics[width=\textwidth]{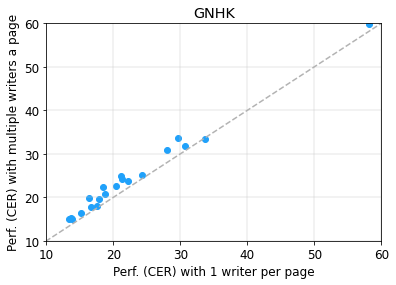}
         \caption{GNHK}
         \label{fig:gnhk_mw}
     \end{subfigure}
     \hfill
     \begin{subfigure}[b]{0.45\textwidth}
         \centering
         \includegraphics[width=\textwidth]{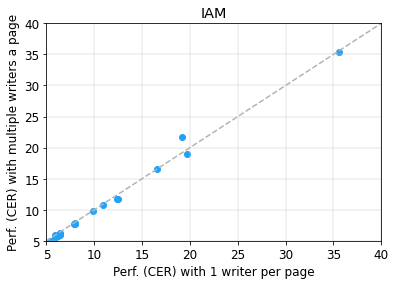}
         \caption{IAM}
         \label{fig:iam_mw}
     \end{subfigure}
     \hfill
     \begin{subfigure}[b]{0.45\textwidth}
         \centering
         \includegraphics[width=\textwidth]{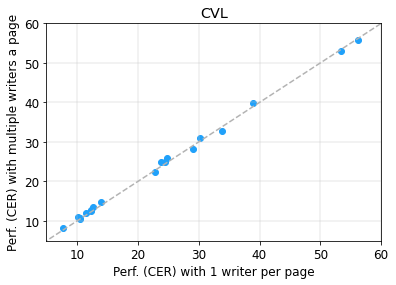}
         \caption{CVL}
         \label{fig:cvl_mw}
     \end{subfigure}
     \hfill
     \begin{subfigure}[b]{0.45\textwidth}
         \centering
         \includegraphics[width=\textwidth]{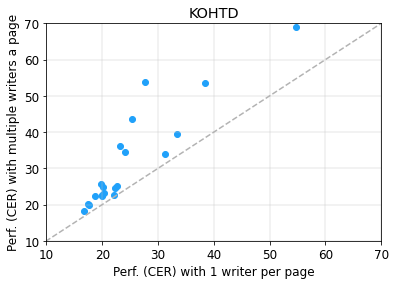}
         \caption{KOHTD}
         \label{fig:kohtd_mw}
     \end{subfigure}
    \caption{\textbf{Multiple writers in a single page.} Each point denotes results for each corrupted version of the particular dataset in the captions. The x-axis of each point is the performance on the vanilla dataset consisting of one writer per page, and the y-axis of that point shows the performance on the concocted version of the dataset consisting multiple writers in a page.}
    \label{fig:multiple_writers}
\end{figure*}

Most of the datasets considered here, and in practice as well, generally consist of only lines from one writer, and thus one writing style. This helps the algorithm to learn from multiple cues across lines, and adapt to the writer's style. This becomes difficult when the same page has writing from multiple writers, as then the number of lines per writer decreases. To do this experiment, we concoct a dataset, where we create pages (i.e., list of lines) by randomly choosing lines from multiple pages (from different writers), without replacement. Then we run our TTA algorithm, except that now it sees lines from multiple writers at a time in a page. We present the performances in Figure \ref{fig:multiple_writers} for all the datasets. The x-axis and y-axis show the performance on the vanilla and concocted dataset. As we can see almost all the points lie above the equi-performance line, indicating that single writer specific adaptation works much better than having multiple writers in a single page. 
%The multiple writer performance even degrades for corruptions where the performance are low, i.e., top-right of the plot.

% \input{./tables_for_supp/multiple_writers}

% \begin{figure}
%     \centering
%     \includegraphics[scale=0.45]{}
%     \label{fig:icdar_mw}
% \end{figure}

% \begin{figure}
%     \centering
%     \includegraphics[scale=0.45]{}
%     \label{fig:gnhk_mw}
% \end{figure}

% \begin{figure}
%     \centering
%     \includegraphics[scale=0.45]{}
%     \label{fig:iam_mw}
% \end{figure}

% Please add the following required packages to your document preamble:
% \usepackage[table,xcdraw]{xcolor}
% If you use beamer only pass "xcolor=table" option, i.e. \documentclass[xcolor=table]{beamer}
\begin{table}
\centering
\resizebox{0.45\textwidth}{!}{
\begin{tabular}{
>{\columncolor[HTML]{FFFFFF}}l 
>{\columncolor[HTML]{FFFFFF}}r 
>{\columncolor[HTML]{FFFFFF}}r 
>{\columncolor[HTML]{FFFFFF}}r 
>{\columncolor[HTML]{FFFFFF}}r 
>{\columncolor[HTML]{FFFFFF}}r 
>{\columncolor[HTML]{FFFFFF}}r 
>{\columncolor[HTML]{FFFFFF}}r }
\toprule
\cellcolor[HTML]{FFFFFF}{\color[HTML]{000000} \textbf{}} &
  \multicolumn{7}{c}{\cellcolor[HTML]{FFFFFF}{\color[HTML]{000000} Number of Iterations}} \\ \midrule
\cellcolor[HTML]{FFFFFF}{\color[HTML]{000000} \textbf{}} &
  \multicolumn{1}{c}{\cellcolor[HTML]{FFFFFF}{\color[HTML]{000000} 1}} &
  \multicolumn{1}{c}{\cellcolor[HTML]{FFFFFF}{\color[HTML]{000000} 2}} &
  \multicolumn{1}{c}{\cellcolor[HTML]{FFFFFF}{\color[HTML]{000000} 4}} &
  \multicolumn{1}{c}{\cellcolor[HTML]{FFFFFF}{\color[HTML]{000000} 6}} &
  \multicolumn{1}{c}{\cellcolor[HTML]{FFFFFF}{\color[HTML]{000000} 8}} &
  \multicolumn{1}{c}{\cellcolor[HTML]{FFFFFF}{\color[HTML]{000000} 10}} &
  \multicolumn{1}{c}{\cellcolor[HTML]{FFFFFF}{\color[HTML]{000000} 12}} \\ \midrule
{\color[HTML]{202124} original} &
  {\color[HTML]{000000} 13.5} &
  {\color[HTML]{000000} \textbf{13.3}} &
  {\color[HTML]{000000} 13.3} &
  {\color[HTML]{000000} 13.4} &
  {\color[HTML]{000000} 13.6} &
  {\color[HTML]{000000} 13.6} &
  {\color[HTML]{000000} 13.8} \\
{\color[HTML]{202124} brightness} &
  {\color[HTML]{000000} 13.4} &
  {\color[HTML]{000000} \textbf{13.2}} &
  {\color[HTML]{000000} 13.2} &
  {\color[HTML]{000000} 13.4} &
  {\color[HTML]{000000} 13.5} &
  {\color[HTML]{000000} 13.5} &
  {\color[HTML]{000000} 13.8} \\
{\color[HTML]{202124} contrast} &
  {\color[HTML]{000000} 15.5} &
  {\color[HTML]{000000} 14.6} &
  {\color[HTML]{000000} \textbf{14.5}} &
  {\color[HTML]{000000} 14.7} &
  {\color[HTML]{000000} 14.8} &
  {\color[HTML]{000000} 14.8} &
  {\color[HTML]{000000} 15.1} \\
{\color[HTML]{202124} defocus blur} &
  {\color[HTML]{000000} 49.8} &
  {\color[HTML]{000000} 43.6} &
  {\color[HTML]{000000} 39.1} &
  {\color[HTML]{000000} 38.4} &
  {\color[HTML]{000000} 38.4} &
  {\color[HTML]{000000} \textbf{38.0}} &
  {\color[HTML]{000000} 38.1} \\
{\color[HTML]{202124} elastic} &
  {\color[HTML]{000000} 48.0} &
  {\color[HTML]{000000} 46.6} &
  {\color[HTML]{000000} 46.3} &
  {\color[HTML]{000000} 43.3} &
  {\color[HTML]{000000} \textbf{43.2}} &
  {\color[HTML]{000000} 43.5} &
  {\color[HTML]{000000} 43.9} \\
{\color[HTML]{202124} fog} &
  {\color[HTML]{000000} 15.8} &
  {\color[HTML]{000000} \textbf{14.9}} &
  {\color[HTML]{000000} 15.1} &
  {\color[HTML]{000000} 15.3} &
  {\color[HTML]{000000} 15.4} &
  {\color[HTML]{000000} 15.5} &
  {\color[HTML]{000000} 15.6} \\
{\color[HTML]{202124} frost} &
  {\color[HTML]{000000} 33.9} &
  {\color[HTML]{000000} 29.0} &
  {\color[HTML]{000000} \textbf{28.6}} &
  {\color[HTML]{000000} 28.9} &
  {\color[HTML]{000000} 29.0} &
  {\color[HTML]{000000} 29.0} &
  {\color[HTML]{000000} 29.5} \\
{\color[HTML]{202124} gaussian blur} &
  {\color[HTML]{000000} 23.2} &
  {\color[HTML]{000000} 18.8} &
  {\color[HTML]{000000} \textbf{18.5}} &
  {\color[HTML]{000000} 18.7} &
  {\color[HTML]{000000} 18.9} &
  {\color[HTML]{000000} 18.7} &
  {\color[HTML]{000000} 19.1} \\
{\color[HTML]{202124} gaussian noise} &
  {\color[HTML]{000000} 24.8} &
  {\color[HTML]{000000} 23.5} &
  {\color[HTML]{000000} \textbf{23.2}} &
  {\color[HTML]{000000} 23.6} &
  {\color[HTML]{000000} 23.7} &
  {\color[HTML]{000000} 23.7} &
  {\color[HTML]{000000} 23.9} \\
{\color[HTML]{202124} glass blur} &
  {\color[HTML]{000000} 37.2} &
  {\color[HTML]{000000} 33.0} &
  {\color[HTML]{000000} 30.5} &
  {\color[HTML]{000000} \textbf{30.4}} &
  {\color[HTML]{000000} 30.5} &
  {\color[HTML]{000000} 30.5} &
  {\color[HTML]{000000} 30.9} \\
{\color[HTML]{202124} impulse noise} &
  {\color[HTML]{000000} 40.5} &
  {\color[HTML]{000000} 34.3} &
  {\color[HTML]{000000} 31.9} &
  {\color[HTML]{000000} \textbf{31.4}} &
  {\color[HTML]{000000} 31.9} &
  {\color[HTML]{000000} 32.0} &
  {\color[HTML]{000000} 32.7} \\
{\color[HTML]{202124} jpeg} &
  {\color[HTML]{000000} 25.7} &
  {\color[HTML]{000000} 23.9} &
  {\color[HTML]{000000} \textbf{23.1}} &
  {\color[HTML]{000000} 23.1} &
  {\color[HTML]{000000} 23.1} &
  {\color[HTML]{000000} 23.1} &
  {\color[HTML]{000000} 23.2} \\
{\color[HTML]{202124} motion blur} &
  {\color[HTML]{000000} 34.2} &
  {\color[HTML]{000000} 28.5} &
  {\color[HTML]{000000} \textbf{28.1}} &
  {\color[HTML]{000000} 28.3} &
  {\color[HTML]{000000} 29.5} &
  {\color[HTML]{000000} 28.2} &
  {\color[HTML]{000000} 28.4} \\
{\color[HTML]{202124} pixelate} &
  {\color[HTML]{000000} 24.9} &
  {\color[HTML]{000000} 21.5} &
  {\color[HTML]{000000} \textbf{20.8}} &
  {\color[HTML]{000000} 21.0} &
  {\color[HTML]{000000} 21.1} &
  {\color[HTML]{000000} 21.2} &
  {\color[HTML]{000000} 23.8} \\
{\color[HTML]{202124} saturate} &
  {\color[HTML]{000000} 13.5} &
  {\color[HTML]{000000} 13.3} &
  {\color[HTML]{000000} \textbf{13.3}} &
  {\color[HTML]{000000} 13.4} &
  {\color[HTML]{000000} 13.5} &
  {\color[HTML]{000000} 13.6} &
  {\color[HTML]{000000} 13.8} \\
{\color[HTML]{202124} shot noise} &
  {\color[HTML]{000000} 36.7} &
  {\color[HTML]{000000} 33.0} &
  {\color[HTML]{000000} 32.0} &
  {\color[HTML]{000000} \textbf{31.8}} &
  {\color[HTML]{000000} 32.1} &
  {\color[HTML]{000000} 32.2} &
  {\color[HTML]{000000} 32.1} \\
{\color[HTML]{202124} snow} &
  {\color[HTML]{000000} 23.4} &
  {\color[HTML]{000000} 21.8} &
  {\color[HTML]{000000} \textbf{21.6}} &
  {\color[HTML]{000000} 21.9} &
  {\color[HTML]{000000} 22.2} &
  {\color[HTML]{000000} 22.1} &
  {\color[HTML]{000000} 22.5} \\
{\color[HTML]{202124} spatter} &
  {\color[HTML]{000000} 15.1} &
  {\color[HTML]{000000} 14.8} &
  {\color[HTML]{000000} \textbf{14.8}} &
  {\color[HTML]{000000} 15.0} &
  {\color[HTML]{000000} 15.1} &
  {\color[HTML]{000000} 15.1} &
  {\color[HTML]{000000} 15.3} \\
{\color[HTML]{202124} speckle noise} &
  {\color[HTML]{000000} 39.7} &
  {\color[HTML]{000000} 35.4} &
  {\color[HTML]{000000} 33.9} &
  {\color[HTML]{000000} \textbf{33.7}} &
  {\color[HTML]{000000} 33.8} &
  {\color[HTML]{000000} 34.1} &
  {\color[HTML]{000000} 34.1} \\
{\color[HTML]{202124} zoom blur} &
  {\color[HTML]{000000} 47.3} &
  {\color[HTML]{000000} 40.1} &
  {\color[HTML]{000000} 36.9} &
  {\color[HTML]{000000} \textbf{36.1}} &
  {\color[HTML]{000000} 36.5} &
  {\color[HTML]{000000} 36.8} &
  {\color[HTML]{000000} 36.5} \\ \midrule
\cellcolor[HTML]{FFFFFF}{\color[HTML]{000000} AVERAGE} &
  \cellcolor[HTML]{FFFFFF}{\color[HTML]{000000} 28.8} &
  \cellcolor[HTML]{FFFFFF}{\color[HTML]{000000} 25.9} &
  \cellcolor[HTML]{FFFFFF}{\color[HTML]{000000} \textbf{24.9}} &
  \cellcolor[HTML]{FFFFFF}{\color[HTML]{000000} \textbf{24.8}} &
  \cellcolor[HTML]{FFFFFF}{\color[HTML]{000000} 25.0} &
  \cellcolor[HTML]{FFFFFF}{\color[HTML]{000000} 25.0} &
  \cellcolor[HTML]{FFFFFF}{\color[HTML]{000000} 25.3} \\ \hline
\end{tabular}
}
\caption{Performance variation (on ICDAR2015-HTR) for different number of progressive updates. For Number of Iterations = $K$, the algorithm is run for $K$ iterations, with $\frac{100}{K}k\%$ of top lines included in the self-training algorithm for iteration number $k \in [1, K]$. As we }
\label{table:topx_icdar}
\end{table}
% Please add the following required packages to your document preamble:
% \usepackage[table,xcdraw]{xcolor}
% If you use beamer only pass "xcolor=table" option, i.e. \documentclass[xcolor=table]{beamer}
\begin{table}
\centering
\resizebox{0.45\textwidth}{!}{
\begin{tabular}{
>{\columncolor[HTML]{FFFFFF}}l 
>{\columncolor[HTML]{FFFFFF}}r 
>{\columncolor[HTML]{FFFFFF}}r 
>{\columncolor[HTML]{FFFFFF}}r 
>{\columncolor[HTML]{FFFFFF}}r 
>{\columncolor[HTML]{FFFFFF}}r 
>{\columncolor[HTML]{FFFFFF}}r 
>{\columncolor[HTML]{FFFFFF}}r }
\toprule
\cellcolor[HTML]{FFFFFF}{\color[HTML]{000000} \textbf{}} &
  \multicolumn{7}{c}{\cellcolor[HTML]{FFFFFF}{\color[HTML]{000000} Number of Iterations}} \\ \midrule
\cellcolor[HTML]{FFFFFF}{\color[HTML]{000000} \textbf{}} &
  \multicolumn{1}{c}{\cellcolor[HTML]{FFFFFF}{\color[HTML]{000000} 1}} &
  \multicolumn{1}{c}{\cellcolor[HTML]{FFFFFF}{\color[HTML]{000000} 2}} &
  \multicolumn{1}{c}{\cellcolor[HTML]{FFFFFF}{\color[HTML]{000000} 4}} &
  \multicolumn{1}{c}{\cellcolor[HTML]{FFFFFF}{\color[HTML]{000000} 6}} &
  \multicolumn{1}{c}{\cellcolor[HTML]{FFFFFF}{\color[HTML]{000000} 8}} &
  \multicolumn{1}{c}{\cellcolor[HTML]{FFFFFF}{\color[HTML]{000000} 10}} &
  \multicolumn{1}{c}{\cellcolor[HTML]{FFFFFF}{\color[HTML]{000000} 12}} \\ \midrule
{\color[HTML]{202124} original} &
  {\color[HTML]{000000} 13.9} &
  {\color[HTML]{000000} 13.9} &
  {\color[HTML]{000000} \textbf{13.7}} &
  {\color[HTML]{000000} 13.8} &
  {\color[HTML]{000000} 13.9} &
  {\color[HTML]{000000} 13.9} &
  {\color[HTML]{000000} 13.9} \\
{\color[HTML]{202124} brightness} &
  {\color[HTML]{000000} 14.2} &
  {\color[HTML]{000000} \textbf{14.1}} &
  {\color[HTML]{000000} 14.2} &
  {\color[HTML]{000000} 14.2} &
  {\color[HTML]{000000} 14.3} &
  {\color[HTML]{000000} 14.3} &
  {\color[HTML]{000000} 14.3} \\
{\color[HTML]{202124} contrast} &
  {\color[HTML]{000000} 15.7} &
  {\color[HTML]{000000} 15.6} &
  {\color[HTML]{000000} \textbf{15.5}} &
  {\color[HTML]{000000} 15.6} &
  {\color[HTML]{000000} 15.6} &
  {\color[HTML]{000000} 15.6} &
  {\color[HTML]{000000} 15.8} \\
{\color[HTML]{202124} defocus blur} &
  {\color[HTML]{000000} 32.7} &
  {\color[HTML]{000000} \textbf{30.9}} &
  {\color[HTML]{000000} 31.0} &
  {\color[HTML]{000000} 31.1} &
  {\color[HTML]{000000} 30.9} &
  {\color[HTML]{000000} 30.9} &
  {\color[HTML]{000000} 30.9} \\
{\color[HTML]{202124} elastic} &
  {\color[HTML]{000000} \textbf{57.5}} &
  {\color[HTML]{000000} 58.2} &
  {\color[HTML]{000000} 58.0} &
  {\color[HTML]{000000} 57.7} &
  {\color[HTML]{000000} 57.8} &
  {\color[HTML]{000000} 57.8} &
  {\color[HTML]{000000} 57.7} \\
{\color[HTML]{202124} fog} &
  {\color[HTML]{000000} 17.0} &
  {\color[HTML]{000000} 17.0} &
  {\color[HTML]{000000} \textbf{16.9}} &
  {\color[HTML]{000000} 17.1} &
  {\color[HTML]{000000} 17.1} &
  {\color[HTML]{000000} 17.3} &
  {\color[HTML]{000000} 17.2} \\
{\color[HTML]{202124} frost} &
  {\color[HTML]{000000} \textbf{20.8}} &
  {\color[HTML]{000000} 20.8} &
  {\color[HTML]{000000} 20.9} &
  {\color[HTML]{000000} 20.9} &
  {\color[HTML]{000000} 21.1} &
  {\color[HTML]{000000} 21.1} &
  {\color[HTML]{000000} 21.2} \\
{\color[HTML]{202124} gaussian blur} &
  {\color[HTML]{000000} 23.6} &
  {\color[HTML]{000000} \textbf{22.6}} &
  {\color[HTML]{000000} 22.7} &
  {\color[HTML]{000000} 22.8} &
  {\color[HTML]{000000} 22.8} &
  {\color[HTML]{000000} 22.8} &
  {\color[HTML]{000000} 22.8} \\
{\color[HTML]{202124} gaussian noise} &
  {\color[HTML]{000000} 19.7} &
  {\color[HTML]{000000} 19.4} &
  {\color[HTML]{000000} \textbf{19.3}} &
  {\color[HTML]{000000} 19.3} &
  {\color[HTML]{000000} 19.4} &
  {\color[HTML]{000000} 19.4} &
  {\color[HTML]{000000} 19.4} \\
{\color[HTML]{202124} glass blur} &
  {\color[HTML]{000000} 37.1} &
  {\color[HTML]{000000} 34.9} &
  {\color[HTML]{000000} \textbf{34.4}} &
  {\color[HTML]{000000} 34.6} &
  {\color[HTML]{000000} 34.7} &
  {\color[HTML]{000000} 34.6} &
  {\color[HTML]{000000} 34.7} \\
{\color[HTML]{202124} impulse noise} &
  {\color[HTML]{000000} 22.1} &
  {\color[HTML]{000000} \textbf{21.6}} &
  {\color[HTML]{000000} 21.7} &
  {\color[HTML]{000000} 21.6} &
  {\color[HTML]{000000} 21.7} &
  {\color[HTML]{000000} 21.7} &
  {\color[HTML]{000000} 21.7} \\
{\color[HTML]{202124} jpeg} &
  {\color[HTML]{000000} 17.3} &
  {\color[HTML]{000000} 17.1} &
  {\color[HTML]{000000} 17.3} &
  {\color[HTML]{000000} 17.2} &
  {\color[HTML]{000000} \textbf{17.0}} &
  {\color[HTML]{000000} 17.3} &
  {\color[HTML]{000000} 17.2} \\
{\color[HTML]{202124} motion blur} &
  {\color[HTML]{000000} 30.4} &
  {\color[HTML]{000000} \textbf{29.8}} &
  {\color[HTML]{000000} 30.1} &
  {\color[HTML]{000000} 30.3} &
  {\color[HTML]{000000} 30.5} &
  {\color[HTML]{000000} 30.7} &
  {\color[HTML]{000000} 30.9} \\
{\color[HTML]{202124} pixelate} &
  {\color[HTML]{000000} 18.8} &
  {\color[HTML]{000000} 18.2} &
  {\color[HTML]{000000} \textbf{18.0}} &
  {\color[HTML]{000000} 18.0} &
  {\color[HTML]{000000} 18.1} &
  {\color[HTML]{000000} 18.3} &
  {\color[HTML]{000000} 18.1} \\
{\color[HTML]{202124} saturate} &
  {\color[HTML]{000000} 14.4} &
  {\color[HTML]{000000} 14.3} &
  {\color[HTML]{000000} \textbf{14.2}} &
  {\color[HTML]{000000} 14.3} &
  {\color[HTML]{000000} 14.3} &
  {\color[HTML]{000000} 14.4} &
  {\color[HTML]{000000} 14.3} \\
{\color[HTML]{202124} shot noise} &
  {\color[HTML]{000000} 22.3} &
  {\color[HTML]{000000} 21.9} &
  {\color[HTML]{000000} 21.8} &
  {\color[HTML]{000000} 21.8} &
  {\color[HTML]{000000} \textbf{21.6}} &
  {\color[HTML]{000000} 21.6} &
  {\color[HTML]{000000} 21.8} \\
{\color[HTML]{202124} snow} &
  {\color[HTML]{000000} 25.0} &
  {\color[HTML]{000000} 25.0} &
  {\color[HTML]{000000} \textbf{25.0}} &
  {\color[HTML]{000000} 25.1} &
  {\color[HTML]{000000} 25.3} &
  {\color[HTML]{000000} 25.3} &
  {\color[HTML]{000000} 25.4} \\
{\color[HTML]{202124} spatter} &
  {\color[HTML]{000000} 18.4} &
  {\color[HTML]{000000} 18.4} &
  {\color[HTML]{000000} \textbf{18.4}} &
  {\color[HTML]{000000} 18.5} &
  {\color[HTML]{000000} 18.5} &
  {\color[HTML]{000000} 18.4} &
  {\color[HTML]{000000} 18.5} \\
{\color[HTML]{202124} speckle noise} &
  {\color[HTML]{000000} 19.3} &
  {\color[HTML]{000000} 19.0} &
  {\color[HTML]{000000} 19.1} &
  {\color[HTML]{000000} 19.0} &
  {\color[HTML]{000000} \textbf{18.9}} &
  {\color[HTML]{000000} 19.0} &
  {\color[HTML]{000000} 19.1} \\
{\color[HTML]{202124} zoom blur} &
  {\color[HTML]{000000} 31.5} &
  {\color[HTML]{000000} 28.7} &
  {\color[HTML]{000000} 28.2} &
  {\color[HTML]{000000} \textbf{28.0}} &
  {\color[HTML]{000000} 28.1} &
  {\color[HTML]{000000} 28.3} &
  {\color[HTML]{000000} 28.1} \\ \midrule
\cellcolor[HTML]{FFFFFF}{\color[HTML]{000000} AVERAGE} &
  \cellcolor[HTML]{FFFFFF}{\color[HTML]{000000} 23.6} &
  \cellcolor[HTML]{FFFFFF}{\color[HTML]{000000} 23.1} &
  \cellcolor[HTML]{FFFFFF}{\color[HTML]{000000} \textbf{23.0}} &
  \cellcolor[HTML]{FFFFFF}{\color[HTML]{000000} \textbf{23.0}} &
  \cellcolor[HTML]{FFFFFF}{\color[HTML]{000000} 23.1} &
  \cellcolor[HTML]{FFFFFF}{\color[HTML]{000000} 23.1} &
  \cellcolor[HTML]{FFFFFF}{\color[HTML]{000000} 23.2} \\ \hline
\end{tabular}
}
\caption{Performance variation (on GNHK) for different number of progressive updates. For Number of Iterations = $K$, the algorithm is run for $K$ iterations, with $\frac{100}{K}k\%$ of top lines included in the self-training algorithm for iteration number $k \in [1, K]$.}
\label{table:topx_gnhk}
\end{table}
% Please add the following required packages to your document preamble:
% \usepackage[table,xcdraw]{xcolor}
% If you use beamer only pass "xcolor=table" option, i.e. \documentclass[xcolor=table]{beamer}
\begin{table}
\centering
\resizebox{0.45\textwidth}{!}{
\begin{tabular}{
>{\columncolor[HTML]{FFFFFF}}l 
>{\columncolor[HTML]{FFFFFF}}r 
>{\columncolor[HTML]{FFFFFF}}r 
>{\columncolor[HTML]{FFFFFF}}r 
>{\columncolor[HTML]{FFFFFF}}r 
>{\columncolor[HTML]{FFFFFF}}r 
>{\columncolor[HTML]{FFFFFF}}r 
>{\columncolor[HTML]{FFFFFF}}r }
\toprule
\cellcolor[HTML]{FFFFFF}{\color[HTML]{000000} \textbf{}} &
  \multicolumn{7}{c}{\cellcolor[HTML]{FFFFFF}{\color[HTML]{000000} Number of Iterations}} \\ \midrule
\cellcolor[HTML]{FFFFFF}{\color[HTML]{000000} \textbf{}} &
  \multicolumn{1}{c}{\cellcolor[HTML]{FFFFFF}{\color[HTML]{000000} 1}} &
  \multicolumn{1}{c}{\cellcolor[HTML]{FFFFFF}{\color[HTML]{000000} 2}} &
  \multicolumn{1}{c}{\cellcolor[HTML]{FFFFFF}{\color[HTML]{000000} 4}} &
  \multicolumn{1}{c}{\cellcolor[HTML]{FFFFFF}{\color[HTML]{000000} 6}} &
  \multicolumn{1}{c}{\cellcolor[HTML]{FFFFFF}{\color[HTML]{000000} 8}} &
  \multicolumn{1}{c}{\cellcolor[HTML]{FFFFFF}{\color[HTML]{000000} 10}} &
  \multicolumn{1}{c}{\cellcolor[HTML]{FFFFFF}{\color[HTML]{000000} 12}} \\ \midrule
{\color[HTML]{202124} original} &
  {\color[HTML]{000000} 5.8}  &
  {\color[HTML]{000000} 5.8} &
  {\color[HTML]{000000} \textbf{5.8}} &
  {\color[HTML]{000000} 5.8} &
  {\color[HTML]{000000} 5.8} &
  {\color[HTML]{000000} 5.8} &
  {\color[HTML]{000000} 5.8} \\
{\color[HTML]{202124} brightness} &
  {\color[HTML]{000000} 7.0} &
  {\color[HTML]{000000} 6.8} &
  {\color[HTML]{000000} \textbf{6.8}} &
  {\color[HTML]{000000} 6.8} &
  {\color[HTML]{000000} 6.8} &
  {\color[HTML]{000000} 6.8} &
  {\color[HTML]{000000} 6.9} \\
{\color[HTML]{202124} contrast} &
  {\color[HTML]{000000} 7.5} &
  {\color[HTML]{000000} \textbf{6.6}} &
  {\color[HTML]{000000} 6.6} &
  {\color[HTML]{000000} 6.6} &
  {\color[HTML]{000000} 6.7} &
  {\color[HTML]{000000} 6.6} &
  {\color[HTML]{000000} 6.7} \\
{\color[HTML]{202124} defocus blur} &
  {\color[HTML]{000000} 24.0} &
  {\color[HTML]{000000} 20.4} &
  {\color[HTML]{000000} \textbf{20.2}} &
  {\color[HTML]{000000} 20.5} &
  {\color[HTML]{000000} 20.6} &
  {\color[HTML]{000000} 20.4} &
  {\color[HTML]{000000} 20.5} \\
{\color[HTML]{202124} elastic} &
  {\color[HTML]{000000} 59.0} &
  {\color[HTML]{000000} 59.1} &
  {\color[HTML]{000000} 58.7} &
  {\color[HTML]{000000} 58.4} &
  {\color[HTML]{000000} 58.3} &
  {\color[HTML]{000000} 58.5} &
  {\color[HTML]{000000} \textbf{58.2}} \\
{\color[HTML]{202124} fog} &
  {\color[HTML]{000000} 6.5} &
  {\color[HTML]{000000} 6.4} &
  {\color[HTML]{000000} \textbf{6.3}} &
  {\color[HTML]{000000} 6.4} &
  {\color[HTML]{000000} 6.4} &
  {\color[HTML]{000000} 6.4} &
  {\color[HTML]{000000} 6.4} \\
{\color[HTML]{202124} frost} &
  {\color[HTML]{000000} 12.5} &
  {\color[HTML]{000000} 11.4} &
  {\color[HTML]{000000} \textbf{11.3}} &
  {\color[HTML]{000000} 11.6} &
  {\color[HTML]{000000} 11.8} &
  {\color[HTML]{000000} 11.7} &
  {\color[HTML]{000000} 11.8} \\
{\color[HTML]{202124} gaussian blur} &
  {\color[HTML]{000000} 14.7} &
  {\color[HTML]{000000} \textbf{12.8}} &
  {\color[HTML]{000000} 12.9} &
  {\color[HTML]{000000} 13.0} &
  {\color[HTML]{000000} 13.1} &
  {\color[HTML]{000000} 13.0} &
  {\color[HTML]{000000} 13.1} \\
{\color[HTML]{202124} gaussian noise} &
  {\color[HTML]{000000} 7.0} &
  {\color[HTML]{000000} 6.9} &
  {\color[HTML]{000000} \textbf{6.8}} &
  {\color[HTML]{000000} 6.9} &
  {\color[HTML]{000000} 6.9} &
  {\color[HTML]{000000} 6.9} &
  {\color[HTML]{000000} 6.9} \\
{\color[HTML]{202124} glass blur} &
  {\color[HTML]{000000} 26.9} &
  {\color[HTML]{000000} 21.6} &
  {\color[HTML]{000000} \textbf{19.7}} &
  {\color[HTML]{000000} 19.8} &
  {\color[HTML]{000000} 19.8} &
  {\color[HTML]{000000} 19.7} &
  {\color[HTML]{000000} 19.9} \\
{\color[HTML]{202124} impulse noise} &
  {\color[HTML]{000000} 10.7} &
  {\color[HTML]{000000} 10.5} &
  {\color[HTML]{000000} \textbf{10.5}} &
  {\color[HTML]{000000} 10.6} &
  {\color[HTML]{000000} 10.7} &
  {\color[HTML]{000000} 10.6} &
  {\color[HTML]{000000} 10.8} \\
{\color[HTML]{202124} jpeg} &
  {\color[HTML]{000000} 7.0} &
  {\color[HTML]{000000} 6.9} &
  {\color[HTML]{000000} 6.9} &
  {\color[HTML]{000000} 6.9} &
  {\color[HTML]{000000} 6.9} &
  {\color[HTML]{000000} \textbf{6.8}} &
  {\color[HTML]{000000} 6.9} \\
{\color[HTML]{202124} motion blur} &
  {\color[HTML]{000000} 19.3} &
  {\color[HTML]{000000} 17.3} &
  {\color[HTML]{000000} \textbf{17.1}} &
  {\color[HTML]{000000} 17.4} &
  {\color[HTML]{000000} 17.5} &
  {\color[HTML]{000000} 17.6} &
  {\color[HTML]{000000} 17.7} \\
{\color[HTML]{202124} pixelate} &
  {\color[HTML]{000000} 9.0} &
  {\color[HTML]{000000} 8.6} &
  {\color[HTML]{000000} \textbf{8.5}} &
  {\color[HTML]{000000} 8.6} &
  {\color[HTML]{000000} 8.5} &
  {\color[HTML]{000000} 8.5} &
  {\color[HTML]{000000} 8.5} \\
{\color[HTML]{202124} saturate} &
  {\color[HTML]{000000} 5.9} &
  {\color[HTML]{000000} 5.9} &
  {\color[HTML]{000000} \textbf{5.8}} &
  {\color[HTML]{000000} 5.8} &
  {\color[HTML]{000000} 5.8} &
  {\color[HTML]{000000} 5.8} &
  {\color[HTML]{000000} 5.9} \\
{\color[HTML]{202124} shot noise} &
  {\color[HTML]{000000} 8.8} &
  {\color[HTML]{000000} 8.7} &
  {\color[HTML]{000000} \textbf{8.6}} &
  {\color[HTML]{000000} 8.7} &
  {\color[HTML]{000000} 8.7} &
  {\color[HTML]{000000} 8.6} &
  {\color[HTML]{000000} 8.7} \\
{\color[HTML]{202124} snow} &
  {\color[HTML]{000000} 13.9} &
  {\color[HTML]{000000} 13.2} &
  {\color[HTML]{000000} \textbf{13.1}} &
  {\color[HTML]{000000} 13.2} &
  {\color[HTML]{000000} 13.2} &
  {\color[HTML]{000000} 13.2} &
  {\color[HTML]{000000} 13.3} \\
{\color[HTML]{202124} spatter} &
  {\color[HTML]{000000} 6.5} &
  {\color[HTML]{000000} 6.4} &
  {\color[HTML]{000000} \textbf{6.4}} &
  {\color[HTML]{000000} 6.4} &
  {\color[HTML]{000000} 6.4} &
  {\color[HTML]{000000} 6.4} &
  {\color[HTML]{000000} 6.4} \\
{\color[HTML]{202124} speckle noise} &
  {\color[HTML]{000000} 8.6} &
  {\color[HTML]{000000} 8.5} &
  {\color[HTML]{000000} \textbf{8.4}} &
  {\color[HTML]{000000} 8.5} &
  {\color[HTML]{000000} 8.5} &
  {\color[HTML]{000000} 8.5} &
  {\color[HTML]{000000} 8.5} \\
{\color[HTML]{202124} zoom blur} &
  {\color[HTML]{000000} 54.1} &
  {\color[HTML]{000000} 42.3} &
  {\color[HTML]{000000} 35.4} &
  {\color[HTML]{000000} 34.5} &
  {\color[HTML]{000000} \textbf{34.3}} &
  {\color[HTML]{000000} 34.9} &
  {\color[HTML]{000000} 35.3} \\ \midrule
\cellcolor[HTML]{FFFFFF}{\color[HTML]{000000} AVERAGE} &
  \cellcolor[HTML]{FFFFFF}{\color[HTML]{000000} 15.7} &
  \cellcolor[HTML]{FFFFFF}{\color[HTML]{000000} 14.3} &
  \cellcolor[HTML]{FFFFFF}{\color[HTML]{000000} \textbf{13.8}} &
  \cellcolor[HTML]{FFFFFF}{\color[HTML]{000000} \textbf{13.8}} &
  \cellcolor[HTML]{FFFFFF}{\color[HTML]{000000} \textbf{13.8}} &
  \cellcolor[HTML]{FFFFFF}{\color[HTML]{000000} \textbf{13.8}} &
  \cellcolor[HTML]{FFFFFF}{\color[HTML]{000000} 13.9} \\ \hline
\end{tabular}
}
\caption{Performance variation (on IAM) for different number of progressive updates. For Number of Iterations = $K$, the algorithm is run for $K$ iterations, with $\frac{100}{K}k\%$ of top lines included in the self-training algorithm for iteration number $k \in [1, K]$.}
% CER for different values of top-k for IAM dataset. Column \textbf{n} indicates 100/n \% of examples in top-k. So, \textbf{n}=2 indicates 100/2, i.e., 50\% of total lines considered as pseudo-labels initially.}
\label{table:topx_iam}
\end{table}
% Please add the following required packages to your document preamble:
% \usepackage[table,xcdraw]{xcolor}
% If you use beamer only pass "xcolor=table" option, i.e. \documentclass[xcolor=table]{beamer}
\begin{table}
\centering
\resizebox{0.45\textwidth}{!}{
\begin{tabular}{
>{\columncolor[HTML]{FFFFFF}}l 
>{\columncolor[HTML]{FFFFFF}}r 
>{\columncolor[HTML]{FFFFFF}}r 
>{\columncolor[HTML]{FFFFFF}}r 
>{\columncolor[HTML]{FFFFFF}}r 
>{\columncolor[HTML]{FFFFFF}}r 
>{\columncolor[HTML]{FFFFFF}}r 
>{\columncolor[HTML]{FFFFFF}}r }
\toprule
\cellcolor[HTML]{FFFFFF}{\color[HTML]{000000} \textbf{}} &
  \multicolumn{7}{c}{\cellcolor[HTML]{FFFFFF}{\color[HTML]{000000} Number of Iterations}} \\ \midrule
\cellcolor[HTML]{FFFFFF}{\color[HTML]{000000} \textbf{}} &
  \multicolumn{1}{c}{\cellcolor[HTML]{FFFFFF}{\color[HTML]{000000} 1}} &
  \multicolumn{1}{c}{\cellcolor[HTML]{FFFFFF}{\color[HTML]{000000} 2}} &
  \multicolumn{1}{c}{\cellcolor[HTML]{FFFFFF}{\color[HTML]{000000} 4}} &
  \multicolumn{1}{c}{\cellcolor[HTML]{FFFFFF}{\color[HTML]{000000} 6}} &
  \multicolumn{1}{c}{\cellcolor[HTML]{FFFFFF}{\color[HTML]{000000} 8}} &
  \multicolumn{1}{c}{\cellcolor[HTML]{FFFFFF}{\color[HTML]{000000} 10}} &
  \multicolumn{1}{c}{\cellcolor[HTML]{FFFFFF}{\color[HTML]{000000} 12}} \\ \midrule
{\color[HTML]{202124} original} &
  {\color[HTML]{000000} 8.4} &
  {\color[HTML]{000000} 8.4} &
  {\color[HTML]{000000} 8.4} &
  {\color[HTML]{000000} 8.4} &
  {\color[HTML]{000000} \textbf{8.3}} &
  {\color[HTML]{000000} 8.4} &
  {\color[HTML]{000000} 8.4} \\
{\color[HTML]{202124} brightness} &
  {\color[HTML]{000000} 11.6} &
  {\color[HTML]{000000} 10.9} &
  {\color[HTML]{000000} 10.8} &
  {\color[HTML]{000000} 10.8} &
  {\color[HTML]{000000} \textbf{10.7}} &
  {\color[HTML]{000000} 10.8} &
  {\color[HTML]{000000} 10.8} \\
{\color[HTML]{202124} contrast} &
  {\color[HTML]{000000} 22.4} &
  {\color[HTML]{000000} 13.0} &
  {\color[HTML]{000000} 12.4} &
  {\color[HTML]{000000} 12.3} &
  {\color[HTML]{000000} \textbf{12.2}} &
  {\color[HTML]{000000} 12.2} &
  {\color[HTML]{000000} 12.2} \\
{\color[HTML]{202124} defocus blur} &
  {\color[HTML]{000000} 39.3} &
  {\color[HTML]{000000} 35.0} &
  {\color[HTML]{000000} 34.1} &
  {\color[HTML]{000000} 34.0} &
  {\color[HTML]{000000} \textbf{33.9}} &
  {\color[HTML]{000000} 34.0} &
  {\color[HTML]{000000} 34.0} \\
{\color[HTML]{202124} elastic} &
  {\color[HTML]{000000} 61.2} &
  {\color[HTML]{000000} 62.5} &
  {\color[HTML]{000000} 60.5} &
  {\color[HTML]{000000} 60.3} &
  {\color[HTML]{000000} 60.4} &
  {\color[HTML]{000000} \textbf{60.2}} &
  {\color[HTML]{000000} 60.2} \\
{\color[HTML]{202124} fog} &
  {\color[HTML]{000000} 11.8} &
  {\color[HTML]{000000} \textbf{11.2}} &
  {\color[HTML]{000000} 11.3} &
  {\color[HTML]{000000} 11.3} &
  {\color[HTML]{000000} 11.2} &
  {\color[HTML]{000000} 11.3} &
  {\color[HTML]{000000} 11.3} \\
{\color[HTML]{202124} frost} &
  {\color[HTML]{000000} 44.4} &
  {\color[HTML]{000000} 39.6} &
  {\color[HTML]{000000} 39.2} &
  {\color[HTML]{000000} \textbf{38.9}} &
  {\color[HTML]{000000} 38.9} &
  {\color[HTML]{000000} 39.0} &
  {\color[HTML]{000000} 39.0} \\
{\color[HTML]{202124} gaussian blur} &
  {\color[HTML]{000000} 30.5} &
  {\color[HTML]{000000} 23.7} &
  {\color[HTML]{000000} 23.5} &
  {\color[HTML]{000000} 23.7} &
  {\color[HTML]{000000} \textbf{23.2}} &
  {\color[HTML]{000000} 23.5} &
  {\color[HTML]{000000} 23.5} \\
{\color[HTML]{202124} gaussian noise} &
  {\color[HTML]{000000} 14.8} &
  {\color[HTML]{000000} \textbf{14.5}} &
  {\color[HTML]{000000} 14.8} &
  {\color[HTML]{000000} 14.8} &
  {\color[HTML]{000000} 14.6} &
  {\color[HTML]{000000} 14.7} &
  {\color[HTML]{000000} 14.7} \\
{\color[HTML]{202124} glass blur} &
  {\color[HTML]{000000} 43.8} &
  {\color[HTML]{000000} 32.7} &
  {\color[HTML]{000000} 29.4} &
  {\color[HTML]{000000} 29.3} &
  {\color[HTML]{000000} \textbf{29.1}} &
  {\color[HTML]{000000} 29.2} &
  {\color[HTML]{000000} 29.2} \\
{\color[HTML]{202124} impulse noise} &
  {\color[HTML]{000000} 28.8} &
  {\color[HTML]{000000} 26.0} &
  {\color[HTML]{000000} \textbf{25.6}} &
  {\color[HTML]{000000} 25.6} &
  {\color[HTML]{000000} 25.7} &
  {\color[HTML]{000000} 25.7} &
  {\color[HTML]{000000} 25.7} \\
{\color[HTML]{202124} jpeg} &
  {\color[HTML]{000000} 14.0} &
  {\color[HTML]{000000} 13.1} &
  {\color[HTML]{000000} 13.1} &
  {\color[HTML]{000000} 13.1} &
  {\color[HTML]{000000} \textbf{13.0}} &
  {\color[HTML]{000000} 13.0} &
  {\color[HTML]{000000} 13.0} \\
{\color[HTML]{202124} motion blur} &
  {\color[HTML]{000000} 35.9} &
  {\color[HTML]{000000} 31.0} &
  {\color[HTML]{000000} \textbf{30.8}} &
  {\color[HTML]{000000} 30.9} &
  {\color[HTML]{000000} 30.8} &
  {\color[HTML]{000000} 30.9} &
  {\color[HTML]{000000} 30.9} \\
{\color[HTML]{202124} pixelate} &
  {\color[HTML]{000000} 15.4} &
  {\color[HTML]{000000} 13.5} &
  {\color[HTML]{000000} 13.2} &
  {\color[HTML]{000000} 13.2} &
  {\color[HTML]{000000} 13.1} &
  {\color[HTML]{000000} \textbf{13.0}} &
  {\color[HTML]{000000} 13.0} \\
{\color[HTML]{202124} saturate} &
  {\color[HTML]{000000} 19.0} &
  {\color[HTML]{000000} 13.7} &
  {\color[HTML]{000000} 13.4} &
  {\color[HTML]{000000} 13.6} &
  {\color[HTML]{000000} \textbf{13.3}} &
  {\color[HTML]{000000} 13.4} &
  {\color[HTML]{000000} 13.4} \\
{\color[HTML]{202124} shot noise} &
  {\color[HTML]{000000} 25.4} &
  {\color[HTML]{000000} \textbf{24.6}} &
  {\color[HTML]{000000} 24.7} &
  {\color[HTML]{000000} 24.6} &
  {\color[HTML]{000000} 24.6} &
  {\color[HTML]{000000} 24.6} &
  {\color[HTML]{000000} 24.6} \\
{\color[HTML]{202124} snow} &
  {\color[HTML]{000000} 59.3} &
  {\color[HTML]{000000} 55.4} &
  {\color[HTML]{000000} 53.3} &
  {\color[HTML]{000000} 53.0} &
  {\color[HTML]{000000} \textbf{52.9}} &
  {\color[HTML]{000000} 52.9} &
  {\color[HTML]{000000} 52.9} \\
{\color[HTML]{202124} spatter} &
  {\color[HTML]{000000} 11.4} &
  {\color[HTML]{000000} \textbf{11.2}} &
  {\color[HTML]{000000} 11.4} &
  {\color[HTML]{000000} 11.3} &
  {\color[HTML]{000000} 11.2} &
  {\color[HTML]{000000} 11.3} &
  {\color[HTML]{000000} 11.3} \\
{\color[HTML]{202124} speckle noise} &
  {\color[HTML]{000000} 25.7} &
  {\color[HTML]{000000} \textbf{25.0}} &
  {\color[HTML]{000000} 25.1} &
  {\color[HTML]{000000} 25.1} &
  {\color[HTML]{000000} 25.1} &
  {\color[HTML]{000000} 25.1} &
  {\color[HTML]{000000} 25.1} \\
{\color[HTML]{202124} zoom blur} &
  {\color[HTML]{000000} 61.1} &
  {\color[HTML]{000000} 57.8} &
  {\color[HTML]{000000} 52.8} &
  {\color[HTML]{000000} 51.5} &
  {\color[HTML]{000000} 51.6} &
  {\color[HTML]{000000} \textbf{51.4}} &
  {\color[HTML]{000000} 51.4} \\ \midrule
\cellcolor[HTML]{FFFFFF}{\color[HTML]{000000} AVERAGE} &
  \cellcolor[HTML]{FFFFFF}{\color[HTML]{000000} 29.2} &
  \cellcolor[HTML]{FFFFFF}{\color[HTML]{000000} 26.1} &
  \cellcolor[HTML]{FFFFFF}{\color[HTML]{000000} \textbf{25.4}} &
  \cellcolor[HTML]{FFFFFF}{\color[HTML]{000000} \textbf{25.3}} &
  \cellcolor[HTML]{FFFFFF}{\color[HTML]{000000} \textbf{25.2}} &
  \cellcolor[HTML]{FFFFFF}{\color[HTML]{000000} \textbf{25.2}} &
  \cellcolor[HTML]{FFFFFF}{\color[HTML]{000000} \textbf{25.2}} \\ \hline
\end{tabular}
}
\caption{Performance variation (on CVL) for different number of progressive updates. For Number of Iterations = $K$, the algorithm is run for $K$ iterations, with $\frac{100}{K}k\%$ of top lines included in the self-training algorithm for iteration number $k \in [1, K]$.}
\label{table:topx_cvl}
\end{table}
% Please add the following required packages to your document preamble:
% \usepackage[table,xcdraw]{xcolor}
% If you use beamer only pass "xcolor=table" option, i.e. \documentclass[xcolor=table]{beamer}
\begin{table}
\centering
\resizebox{0.45\textwidth}{!}{
\begin{tabular}{
>{\columncolor[HTML]{FFFFFF}}l 
>{\columncolor[HTML]{FFFFFF}}r 
>{\columncolor[HTML]{FFFFFF}}r 
>{\columncolor[HTML]{FFFFFF}}r 
>{\columncolor[HTML]{FFFFFF}}r 
>{\columncolor[HTML]{FFFFFF}}r 
>{\columncolor[HTML]{FFFFFF}}r 
>{\columncolor[HTML]{FFFFFF}}r }
\toprule
\cellcolor[HTML]{FFFFFF}{\color[HTML]{000000} \textbf{}} &
  \multicolumn{7}{c}{\cellcolor[HTML]{FFFFFF}{\color[HTML]{000000} Number of Iterations}} \\ \midrule
\cellcolor[HTML]{FFFFFF}{\color[HTML]{000000} \textbf{}} &
  \multicolumn{1}{c}{\cellcolor[HTML]{FFFFFF}{\color[HTML]{000000} 1}} &
  \multicolumn{1}{c}{\cellcolor[HTML]{FFFFFF}{\color[HTML]{000000} 2}} &
  \multicolumn{1}{c}{\cellcolor[HTML]{FFFFFF}{\color[HTML]{000000} 4}} &
  \multicolumn{1}{c}{\cellcolor[HTML]{FFFFFF}{\color[HTML]{000000} 6}} &
  \multicolumn{1}{c}{\cellcolor[HTML]{FFFFFF}{\color[HTML]{000000} 8}} &
  \multicolumn{1}{c}{\cellcolor[HTML]{FFFFFF}{\color[HTML]{000000} 10}} &
  \multicolumn{1}{c}{\cellcolor[HTML]{FFFFFF}{\color[HTML]{000000} 12}} \\ \midrule
{\color[HTML]{202124} original} &
  {\color[HTML]{000000} 19.3} &
  {\color[HTML]{000000} 17.3} &
  {\color[HTML]{000000} 16.8} &
  {\color[HTML]{000000} \textbf{16.1}} &
  {\color[HTML]{000000} 16.3} &
  {\color[HTML]{000000} 16.4} &
  {\color[HTML]{000000} 16.6} \\
{\color[HTML]{202124} brightness} &
  {\color[HTML]{000000} 23.5} &
  {\color[HTML]{000000} 21.5} &
  {\color[HTML]{000000} 19.9} &
  {\color[HTML]{000000} \textbf{19.7}} &
  {\color[HTML]{000000} 19.8} &
  {\color[HTML]{000000} 19.9} &
  {\color[HTML]{000000} 20.2} \\
{\color[HTML]{202124} contrast} &
  {\color[HTML]{000000} 24.3} &
  {\color[HTML]{000000} 21.9} &
  {\color[HTML]{000000} 20.1} &
  {\color[HTML]{000000} \textbf{19.8}} &
  {\color[HTML]{000000} 19.8} &
  {\color[HTML]{000000} 20.0} &
  {\color[HTML]{000000} 20.3} \\
{\color[HTML]{202124} defocus blur} &
  {\color[HTML]{000000} 37.5} &
  {\color[HTML]{000000} 35.6} &
  {\color[HTML]{000000} 33.4} &
  {\color[HTML]{000000} \textbf{32.6}} &
  {\color[HTML]{000000} 32.6} &
  {\color[HTML]{000000} 32.6} &
  {\color[HTML]{000000} 32.9} \\
{\color[HTML]{202124} elastic} &
  {\color[HTML]{000000} 25.9} &
  {\color[HTML]{000000} 23.9} &
  {\color[HTML]{000000} 22.3} &
  {\color[HTML]{000000} \textbf{22.1}} &
  {\color[HTML]{000000} 22.2} &
  {\color[HTML]{000000} 22.4} &
  {\color[HTML]{000000} 22.6} \\
{\color[HTML]{202124} fog} &
  {\color[HTML]{000000} 28.4} &
  {\color[HTML]{000000} 25.8} &
  {\color[HTML]{000000} 24.0} &
  {\color[HTML]{000000} \textbf{23.7}} &
  {\color[HTML]{000000} 23.7} &
  {\color[HTML]{000000} 24.0} &
  {\color[HTML]{000000} 24.0} \\
{\color[HTML]{202124} frost} &
  {\color[HTML]{000000} 56.5} &
  {\color[HTML]{000000} 54.6} &
  {\color[HTML]{000000} 54.8} &
  {\color[HTML]{000000} 54.0} &
  {\color[HTML]{000000} 54.1} &
  {\color[HTML]{000000} 53.9} &
  {\color[HTML]{000000} \textbf{53.8}} \\
{\color[HTML]{202124} gaussian blur} &
  {\color[HTML]{000000} 23.6} &
  {\color[HTML]{000000} 21.6} &
  {\color[HTML]{000000} 19.9} &
  {\color[HTML]{000000} \textbf{19.6}} &
  {\color[HTML]{000000} 19.7} &
  {\color[HTML]{000000} 19.8} &
  {\color[HTML]{000000} 20.1} \\
{\color[HTML]{202124} gaussian noise} &
  {\color[HTML]{000000} 23.3} &
  {\color[HTML]{000000} 21.2} &
  {\color[HTML]{000000} 19.7} &
  {\color[HTML]{000000} \textbf{19.5}} &
  {\color[HTML]{000000} 19.6} &
  {\color[HTML]{000000} 19.7} &
  {\color[HTML]{000000} 19.9} \\
{\color[HTML]{202124} glass blur} &
  {\color[HTML]{000000} 27.5} &
  {\color[HTML]{000000} 24.7} &
  {\color[HTML]{000000} 22.7} &
  {\color[HTML]{000000} \textbf{22.5}} &
  {\color[HTML]{000000} 22.5} &
  {\color[HTML]{000000} 22.8} &
  {\color[HTML]{000000} 23.0} \\
{\color[HTML]{202124} impulse noise} &
  {\color[HTML]{000000} 31.1} &
  {\color[HTML]{000000} 29.1} &
  {\color[HTML]{000000} 27.6} &
  {\color[HTML]{000000} \textbf{27.5}} &
  {\color[HTML]{000000} 27.6} &
  {\color[HTML]{000000} 27.6} &
  {\color[HTML]{000000} 27.8} \\
{\color[HTML]{202124} jpeg} &
  {\color[HTML]{000000} 24.6} &
  {\color[HTML]{000000} 22.2} &
  {\color[HTML]{000000} 20.3} &
  {\color[HTML]{000000} \textbf{20.1}} &
  {\color[HTML]{000000} 20.2} &
  {\color[HTML]{000000} 20.3} &
  {\color[HTML]{000000} 20.5} \\
{\color[HTML]{202124} motion blur} &
  {\color[HTML]{000000} 34.7} &
  {\color[HTML]{000000} 33.0} &
  {\color[HTML]{000000} 31.3} &
  {\color[HTML]{000000} \textbf{30.8}} &
  {\color[HTML]{000000} 30.8} &
  {\color[HTML]{000000} 31.0} &
  {\color[HTML]{000000} 31.1} \\
{\color[HTML]{202124} pixelate} &
  {\color[HTML]{000000} 22.8} &
  {\color[HTML]{000000} 20.4} &
  {\color[HTML]{000000} \textbf{18.7}} &
  {\color[HTML]{000000} 18.7} &
  {\color[HTML]{000000} 18.8} &
  {\color[HTML]{000000} 18.8} &
  {\color[HTML]{000000} 19.1} \\
{\color[HTML]{202124} saturate} &
  {\color[HTML]{000000} 21.0} &
  {\color[HTML]{000000} 18.9} &
  {\color[HTML]{000000} 17.4} &
  {\color[HTML]{000000} \textbf{17.3}} &
  {\color[HTML]{000000} 17.5} &
  {\color[HTML]{000000} 17.5} &
  {\color[HTML]{000000} 17.8} \\
{\color[HTML]{202124} shot noise} &
  {\color[HTML]{000000} 26.7} &
  {\color[HTML]{000000} 24.7} &
  {\color[HTML]{000000} 23.1} &
  {\color[HTML]{000000} \textbf{23.0}} &
  {\color[HTML]{000000} 23.1} &
  {\color[HTML]{000000} 23.2} &
  {\color[HTML]{000000} 23.4} \\
{\color[HTML]{202124} snow} &
  {\color[HTML]{000000} 41.7} &
  {\color[HTML]{000000} 39.4} &
  {\color[HTML]{000000} \textbf{38.5}} &
  {\color[HTML]{000000} 38.6} &
  {\color[HTML]{000000} 38.7} &
  {\color[HTML]{000000} 39.0} &
  {\color[HTML]{000000} 39.0} \\
{\color[HTML]{202124} spatter} &
  {\color[HTML]{000000} 21.2} &
  {\color[HTML]{000000} 19.2} &
  {\color[HTML]{000000} 17.7} &
  {\color[HTML]{000000} \textbf{17.6}} &
  {\color[HTML]{000000} 17.8} &
  {\color[HTML]{000000} 17.9} &
  {\color[HTML]{000000} 18.1} \\
{\color[HTML]{202124} speckle noise} &
  {\color[HTML]{000000} 28.7} &
  {\color[HTML]{000000} 26.8} &
  {\color[HTML]{000000} 25.3} &
  {\color[HTML]{000000} \textbf{24.9}} &
  {\color[HTML]{000000} 25.2} &
  {\color[HTML]{000000} 25.3} &
  {\color[HTML]{000000} 25.4} \\
{\color[HTML]{202124} zoom blur} &
  {\color[HTML]{000000} 26.3} &
  {\color[HTML]{000000} 24.0} &
  {\color[HTML]{000000} 22.1} &
  {\color[HTML]{000000} \textbf{21.7}} &
  {\color[HTML]{000000} 21.8} &
  {\color[HTML]{000000} 21.8} &
  {\color[HTML]{000000} 22.1} \\ \midrule
\cellcolor[HTML]{FFFFFF}{\color[HTML]{000000} AVERAGE} &
  \cellcolor[HTML]{FFFFFF}{\color[HTML]{000000} 28.4} &
  \cellcolor[HTML]{FFFFFF}{\color[HTML]{000000} 26.3} &
  \cellcolor[HTML]{FFFFFF}{\color[HTML]{000000} \textbf{24.8}} &
  \cellcolor[HTML]{FFFFFF}{\color[HTML]{000000} \textbf{24.5}} &
  \cellcolor[HTML]{FFFFFF}{\color[HTML]{000000} \textbf{24.6}} &
  \cellcolor[HTML]{FFFFFF}{\color[HTML]{000000} \textbf{24.7}} &
  \cellcolor[HTML]{FFFFFF}{\color[HTML]{000000} \textbf{24.9}} \\ \hline
\end{tabular}
}
\caption{Performance variation (on KOHTD) for different number of progressive updates. For Number of Iterations = $K$, the algorithm is run for $K$ iterations, with $\frac{100}{K}k\%$ of top lines included in the self-training algorithm for iteration number $k \in [1, K]$.}
\label{table:topx_kohtd}
\end{table}

\end{document}